\documentclass[10pt,journal,compsoc]{IEEEtran}

\usepackage{pifont}
\newcommand{\xmark}{\ding{55}}%
\usepackage{multirow}

\usepackage{amssymb}

\usepackage{amsmath}
\usepackage{ragged2e}
\usepackage{float}
\ifCLASSOPTIONcompsoc

\usepackage[nocompress]{cite}
\else

\usepackage{cite}
\fi

\usepackage{url} 

\usepackage{times}
\usepackage{graphicx} 
\usepackage{subfigure} 

\usepackage[ruled,linesnumbered]{algorithm2e}

\usepackage{hyperref}

\usepackage{helvet}
\usepackage{courier}

\usepackage{makecell}
\usepackage{graphicx}
\usepackage{subfigure}
\usepackage{color}
\usepackage{amsfonts}
\usepackage{amsmath}
\usepackage{amssymb}

\usepackage{multirow}
\usepackage{booktabs}

\DeclareMathOperator*{\argmin}{arg\,min}
\DeclareMathOperator*{\argmax}{arg\,max}

\def\etal{{\em et al.\/}\, }

\def\mB{{\mathcal B}}

\def\mH{{\mathcal H}}

\def\mL{{\mathcal L}}

\def\mS{{\mathcal S}}

\def\mV{{\mathcal V}}

\def\mX{{\mathcal X}}
\def\mY{{\mathcal Y}}
\def\mZ{{\mathcal{Z}}}

\DeclareMathAlphabet\mathbfcal{OMS}{cmsy}{b}{n}

\def\0{{\bf 0}}
\def\1{{\bf 1}}

\def\bW{{\bf W}}
\def\bX{{\bf X}}

\def\bx{{\bf x}}
\def\by{{\bf y}}

\def\bx{{\bf x}}
\def\bX{{\bf X}}
\def\by{{\bf y}}

\def\bW{{\bf W}}

\def\ie{\mbox{{i.e.}}}
\def\eg{\mbox{{e.g.}}}
\def\wrt{\mbox{{w.r.t. }}}

\usepackage{ntheorem}

\newtheorem{deftn}{Definition}
\newtheorem{thm}{Theorem}
\newtheorem*{*thm}{Theorem}

\newtheorem*{*lemma}{Lemma}

\usepackage{enumitem}

\def\qi{\textcolor{black}}
\definecolor{dzs}{RGB}{100, 20, 200} 
\def\dzs{\textcolor{black}}

\definecolor{AJEcolor}{RGB}{60, 179, 113}

\newcommand{\eat}[1]{}
  
\usepackage{xspace}

\newcommand{\guo}[1]{\textcolor[rgb]{0.0, 0.0, 0.0}{#1}}
\def\mytitle{Towards Lightweight Super-Resolution with Dual Regression Learning}

\begin{document}
	
	\title{\mytitle}
	\author{
        \small{Yong Guo$^*$, Mingkui Tan$^{*\dagger}$, Zeshuai Deng$^*$, Jingdong Wang, Qi Chen, Jiezhang Cao, Yanwu Xu, Jian Chen$^*$}
	\IEEEcompsocitemizethanks{
	\IEEEcompsocthanksitem{Yong Guo, Mingkui Tan, Zeshuai Deng and Jian Chen are with the School of Software Engineering, South China University of Technology. Yong Guo is also with the Max Planck Institute for Informatics. Mingkui Tan and Zeshuai Deng are also with the Pazhou Laboratory, Guangzhou, China.
    E-mail: guoyongcs@gmail.com, 
    \{mingkuitan, ellachen\}@scut.edu.cn},
    sedengzeshuai@mail.scut.edu.cn
	\IEEEcompsocthanksitem{Jingdong Wang is with the Baidu Inc. E-mail: wangjingdong@baidu.com}
	\IEEEcompsocthanksitem{Qi Chen is with the Faculty of Engineering, the University of Adelaide. E-mail: qi.chen04@adelaide.edu.au}
	\IEEEcompsocthanksitem{Jiezhang Cao is with ETH Zürich. E-mail: jiezhang.cao@vision.ee.ethz.ch}
        \IEEEcompsocthanksitem{Yanwu Xu is with the School of Future Technology, South China University of Technology. E-mail: ywxu@ieee.org}
	\IEEEcompsocthanksitem{
	$^*$ Authors contributed equally. 
	$^{\dagger}$ Corresponding author.}
	}
    }

	\markboth{Journal of \LaTeX\ Class Files, 2024}%
	{Shell \MakeLowercase{\textit{et al.}}: \mytitle}
	
	\IEEEtitleabstractindextext{%
		\begin{abstract}
		\justifying
            Deep neural networks have exhibited remarkable performance in image super-resolution (SR) tasks by learning a mapping from low-resolution (LR) images to high-resolution (HR) images.
            However, the SR problem is typically an ill-posed problem and existing methods would come with several limitations.
            {First, the possible mapping space of SR can be extremely large since there may exist many different HR images that can be super-resolved from the same LR image. As a result, it is hard to directly learn a promising SR mapping from such a large space.}
            Second, it is often inevitable to develop very large models with extremely high computational cost
            to yield promising SR performance. 
            In practice, 
            one can use model compression techniques to obtain compact models by reducing model redundancy. Nevertheless, it is hard for existing model compression methods to accurately identify the redundant components due to the extremely large SR mapping space.
            To alleviate the first challenge, we propose a dual regression learning scheme to reduce the space of possible SR mappings.
            Specifically, in addition to the mapping from LR to HR images, we learn an additional dual regression mapping to estimate the downsampling kernel and reconstruct LR images. In this way, the dual mapping acts as a constraint to reduce the space of possible mappings. 
            To address the second challenge, we propose a dual regression compression (DRC) method to reduce model redundancy in both layer-level and channel-level based on channel pruning. Specifically, we first develop a channel number search method that minimizes the dual regression loss to determine the redundancy of each layer. 
            Given the searched channel numbers, we further exploit the dual regression manner to evaluate the importance of channels and prune the redundant ones.
            Extensive experiments show the effectiveness of our method in obtaining accurate and efficient SR models.
		\end{abstract}
		
    	}

	\maketitle

	\IEEEdisplaynontitleabstractindextext

	\IEEEpeerreviewmaketitle

	\IEEEraisesectionheading{\section{Introduction}\label{sec:introduction}}
	
	\IEEEPARstart{D}{eep} neural networks (DNNs) have been the workhorse of many real-world applications, including image classification~\cite{he2016deep, russakovsky2015imagenet}
	and image restoration~\cite{wang2015deep, dong2018denoising, zhang2020residual, deng2020deep, lai2018fast, zhang2019ranksrgan, hu2019meta, anwar2020densely, zhou2020cross, kong2021classsr, liang2021flow, gu2020interpreting, zhang2022fluid, zhang2022heat, zhang2024mdeformer, deng2024efficient}.
    Recently, image super-resolution (SR) has become an important task
    that aims to learn a non-linear mapping to reconstruct high-resolution (HR) images from low-resolution (LR) images. 
    Nevertheless, the SR problem is typically an ill-posed problem and it is non-trivial
    to learn an effective SR model due to several underlying challenges.

    First, the space of possible SR mapping functions can be extremely large since there exist many HR images that can be super-resolved from the same LR image~\cite{ulyanov2018deep}.
    \guo{
    In practice, most methods directly optimize the reconstruction loss (\eg, MAE or MSE) in HR domain and often easily obtain very blurry results with insufficient high-frequency information~\cite{wang2018recovering}. In other words, these undesired blurry solutions may take up the majority of the space of SR mapping to be learned and make the whole learning space extremely large.   
    As a result, it is non-trivial to find a promising one from such a large space.}
    To alleviate this issue, existing methods seek to increase the model capacity (\eg, EDSR~\cite{lim2017enhanced}, RCAN~\cite{zhang2018image} and ESSAformer~\cite{zhang2023essaformer}) and minimize the reconstruction error between the super-resolved images and the ground-truth HR images.
    However, these methods still suffer from such a large space of possible SR mapping functions (See analysis in Section~\ref{sec:dual_scheme}) and often yield limited performance.
    Thus, how to reduce the possible space of the mapping functions to boost the training of SR models becomes an important problem.
    
    Second, most SR models often contain a large number of parameters and come with extremely high computational cost. 
    To address this, many efforts have been made to design efficient SR models~\cite{dong2016accelerating, hui2018fast}.
    However, these models often incur a dramatic performance gap compared with state-of-the-art SR methods~\cite{mei2020image,niu2020single}.
    Unlike these methods, one can also exploit model compression techniques (\eg, channel pruning) to obtain lightweight models.
    Nevertheless, it is non-trivial to identify the redundant components (\eg, channels) in SR models due to the large possible mapping space.
    Specifically, given an inaccurate SR mapping, the estimated/predicted redundancy of model components may be also very inaccurate.
    More critically, the redundancy may vary a lot among different layers in the model and different channels in each layer, making it harder to identify the redundant components.

    In this paper, we propose a novel dual regression learning scheme to obtain accurate and efficient SR models.
    {To reduce the possible mapping space, we introduce an additional constraint that encourages the super-resolved images to reconstruct the input LR images.}
    \guo{Suppose there are some high-frequency textures (\eg, contour of object or human hair) inside LR images, this constraint guarantees that the super-resolved images are able to preserve the high-frequency information if they can perfectly reconstruct the original LR images.
    In other words, we are able to effectively exclude a large number of solutions that catastrophically lose these high-frequency textures even though they perform very well on the low-frequency parts (often with a very small loss w.r.t. MAE or MSE).
    }
    With this constraint, the dual regression scheme improves SR performance by reducing the space of possible SR mappings, yielding a smaller generalization bound than existing methods (See Theorem~\ref{theorem:generalization_bound}). 
    {To obtain effective lightweight SR models, we propose a search-guided pruning pipeline, named the dual regression compression (DRC) method, to reduce the model redundancy in both layer-level and channel-level.
    Specifically, we first determine the redundancy of each layer by performing channel number search with our dual regression scheme.
    Unlike existing methods,
    we design an importance-aware search strategy to facilitate the channel number search for pruning.
    Then, we exploit the dual regression scheme to evaluate the importance of channels and prune those redundant ones according to the searched channel numbers. 
    Extensive experiments under both the non-blind and blind SR settings demonstrate the superiority of our method (See results in Table~\ref{tab:4xsr}, \ref{tab:lightweight_comparison} and \ref{tab:compare_blind}).
    }

    Our contributions are summarized as follows:
    \begin{itemize}[leftmargin=*]

    \item 
    To alleviate the issue of extremely large SR mapping space incurred by the nature of ill-posed problems, we propose a dual regression learning scheme that introduces an additional dual mapping to reconstruct LR images. The dual mapping acts as a constraint to reduce the space of possible SR mapping functions and enhance the training of SR models.

        \item  {
        Unlike most model compression methods, we propose a search-guided pruning pipeline, dual regression compression method (DRC), to exploit a reduced mapping space to identify both the layer-level and channel-level redundancy. 
        Specifically, we design an importance-aware search strategy to identify the redundancy of each layer and search for a promising channel configuration for the subsequent pruning process. Then, we conduct channel pruning to remove the redundant channels according to the searched channel number configuration.
        }


     \item {
    Extensive experiments demonstrate the flexibility of our dual regression scheme for SR.
    In practice, our method is applicable to boost the training and model compression for both CNN-based and transformer-based SR models. Furthermore, we demonstrate that our DRC scheme is able to achieve a lossless compression to reduce the computational cost under both the non-blind and the blind SR settings.
    }
    \end{itemize}

    This paper extends our preliminary version~\cite{guo2020dual} from several aspects. 
    {
    1) We propose a dual regression compression scheme (DRC) to achieve lossless compression for SR models. Unlike most existing methods, our DRC simultaneously conduct channel number search and channel pruning to enhance the performance of model compression.
    2) We present a dual regression based channel number search method to identify the layer-level redundancy by determining the number of channels for each layer. During the search process, we design an importance-aware search strategy to facilitate the channel number search for pruning.
    3) We develop a dual regression based channel pruning algorithm that exploits the dual regression manner to evaluate the importance of channels when performing channel pruning.
    4) We conduct more experiments on both CNN-based and transformer-based SR models to investigate the effect of our dual regression scheme.
    5) We demonstrate the effectiveness of our dual regression compression scheme under both the non-blind and blind SR settings.
    }
     
	\section{Related Work} \label{sec:related_work}
	
	\subsection{Image Super-resolution}
	
    Existing SR methods mainly include interpolation-based approaches~\cite{hou1978cubic, allebach1996edge, li2001new, nguyen2000efficient} and reconstruction-based methods~\cite{DBPN2018, li2019feedback, zhang2018image, shocher2018zero, gu2019blind}.
    Interpolation-based methods may oversimplify the SR problem and usually generate blurry images~\cite{ledig2017photo,tong2017image}.
    The reconstruction-based methods~\cite{baker2002limits,ben2007penrose,lin2004fundamental} reconstruct the HR images from LR images.
    Following such methods, many CNN-based methods~\cite{dong2014learning, dong2015image, kim2016accurate, mao2016image, zhang2018learning, zhang2018residual,guo2020hierarchical, zhang2022curvature} were developed to learn a reconstruction mapping.
    
    Recently, 
    Ledig~\etal\cite{ledig2017photo} propose a deep residual network SRResNet for super-resolution. Lim~\etal\cite{lim2017enhanced} remove unnecessary modules in the residual network~\cite{he2016deep} and design a very wide network EDSR.
    Haris~\etal\cite{DBPN2018} propose a back-projection network (DBPN) to iteratively produce LR and HR images. 
    Zhang~\etal\cite{zhang2018image} propose a channel attention mechanism to build a deep model called RCAN to further improve the SR performance.
    Mei~\etal\cite{mei2020image} propose a Cross-Scale Non-Local attention module for more accurate image SR.
    Niu~\etal\cite{niu2020single} propose a holistic attention network (HAN) to model the interdependencies among layers, channels, and spatial positions. 
    Liang~\etal \cite{liang2021swinir} develop a transformer model to improve the performance of image restoration.
    {Chen~\etal \cite{chen2023dual} further propose a dual aggregation transformer to effectively aggregate features across spatial and channel dimensions. Moreover, Zhang~\etal~\cite{zhang2022pixel} focus on blind deblurring and develop a pixel screening based intermediate correction method.}
    However, the training process of these methods still has a very large space of the possible SR mappings, making it hard to learn a good solution in practice.

	\begin{figure}[t]
	\centering
	\includegraphics[width=1\columnwidth]{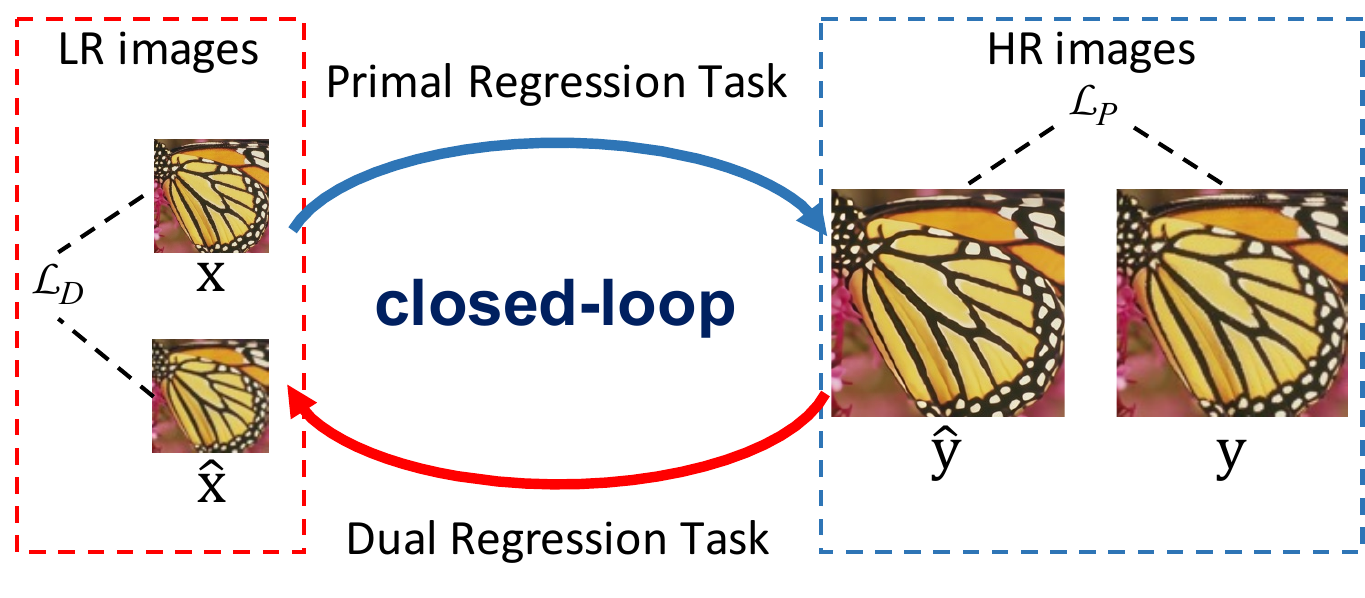}
	\caption{
		The proposed dual regression learning scheme contains a primal regression task for SR and a dual regression task to reconstruct LR images. The primal and dual regression tasks form a closed-loop.
	}
	\label{fig:dual_connection}
    \end{figure}

    \begin{figure*}[t]
    	\centering
    	\includegraphics[width=0.97\textwidth]{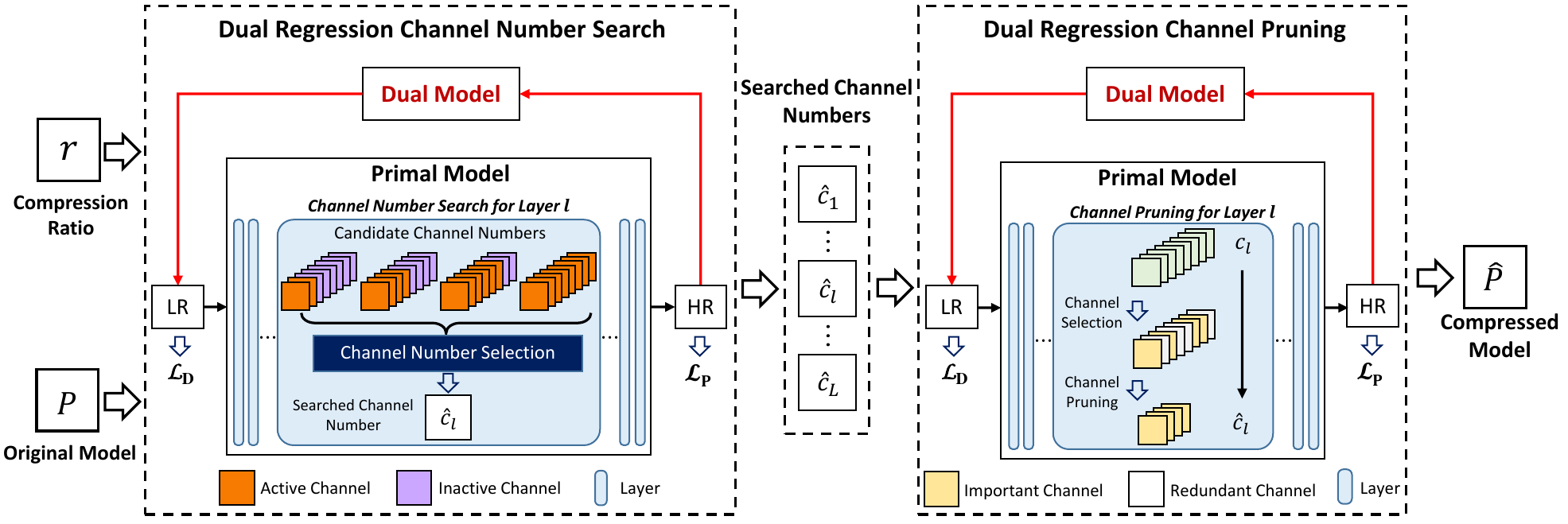}
    	\caption{Overview of the \textbf{dual regression compression (DRC)} approach. 
        {Given a target compression ratio $r$, we first determine the redundancy of each layer by performing the dual regression based channel number search.
        Then, according to the searched channel numbers, we evaluate the importance of channels and prune those redundant ones to obtain the compressed model $\widehat{P}$.}
    	}
    	\label{fig:dual_compression}
    \end{figure*}

	\subsection{Lightweight Model Techniques}
    Lightweight models have gained great attention in recent years~\cite{wang2020deep, li2023ntire, zhang2024irprunedet}. One can obtain lightweight models by directly designing efficient architectures or distilling knowledge from other models. 
    Hui~\etal\cite{hui2018fast} propose an information distillation block to extract the local long and short-path features for lightweight SR networks.
    Zhang~\etal\cite{zhang2021edge} propose a re-parameterizable building block for efficient SR.
    However, these models often incur a dramatic performance gap compared to state-of-the-art SR methods~\cite{mei2020image,niu2020single}. Besides these methods, one can enhance the lightweight SR performance using knowledge distillation technique~\cite{gao2018image, lee2020learning, zhang2021data}. Gao~\etal\cite{gao2018image} use a lightweight student SR model to learn the knowledge from the deeper teacher SR network. 
    Lee~\etal\cite{lee2020learning} propose a distillation framework that leverages HR images as privileged information to boost the training of the student network.
    {
    Wang~\etal \cite{wang2021exploring} explore the sparsity in image SR and predict a pixel-level redundancy mask to improve the inference efficiency of SR networks.
    Yu~\etal\cite{lee2020learning} propose a hyperparameter optimization method to search for an efficient architecture for super-resolution from scratch. Yu~\etal~\cite{yu2021efficient} shares a similar idea with us to search/find the optimal number of channels. 
    However, both methods define a very limited search space where all the blocks/cells share the same number of channels, ignoring the differences of redundancy among different layers. Thus, they cannot identify the layer-wise redundancy of channels and may still result in suboptimal model performance.
    Zhan~\etal \cite{zhan2021achieving} further combine neural architecture search with layer-wise pruning strategy to obtain efficient SR models.
    }
    
    Besides these methods, we can also use model compression techniques to obtain lightweight models~\cite{Han2015NIPS, Han2016Compression, Luo2019ThiNet,guo2021towards}.
    As one of the predominant approaches, channel pruning~\cite{zhuang2018nips, Lin2020HRank, liu2019metapruning, dong2019network, lin2020channel, wang2020revisiting} seeks to remove the redundant channels of deep models to obtain compact subnets. It has been shown that these subnets often come with promising accuracy~\cite{frankle2018lottery} and robustness~\cite{guo2022improving,sehwag2020hydra,li2020towards}.
    Recently, 
    {
    Hou~\etal \cite{hou2020efficient} use conditional covariance~\cite{baker1973joint} to measure the importance of channels to the final output of SR models. 
    }
    Li~\etal\cite{li2020dhp} propose a differentiable meta channel pruning method (DHP) to compress SR models.
    {
    Zhang~\etal\cite{zhang2022learning} impose regularization on the pruned
    structure to ensure the locations of pruned filters are aligned across different layers of residual blocks.
    }
    In addition, some quantization-based methods~\cite{xin2020binarized, ma2019efficient, li2020pams} exploit low bits to accelerate the inference speed of SR models.
    However, it is still non-trivial for these methods to identify the redundant components due to the extremely large possible function space. 
	Unlike them, we seek to reduce the possible function space to alleviate the training/compression difficulty. Thus, it becomes possible to obtain lightweight SR models without significant performance degradation (See Fig.~\ref{tab:lightweight_comparison}).

	\subsection{Dual Learning}
    Dual learning~\cite{he2016dual,xia2017dual,xia2018model,zhang2018deep}
    contains a primal model and a dual model and learns two opposite mappings simultaneously to enhance the performance of language translation.
    Recently, this scheme has also been used to perform image translation without paired training data~\cite{zhu2017unpaired,yi2017dualgan}.
    Specifically, a cycle consistency loss is proposed to avoid the mode collapse issue of GAN methods~\cite{zhu2017unpaired,cao2018adversarial,guo2019auto} and help minimize the distribution divergence. However, these methods cannot be directly applied to the standard SR problem.
    By contrast, we use the closed-loop to reduce the space of possible functions of SR. Moreover, we consider learning asymmetric mappings and provide a theoretical guarantee on the rationality and necessity of using a cycle.

	\section{Dual Regression Networks} \label{sec:dual_regression_learning}
	In this paper, we propose a dual regression learning scheme to obtain accurate and efficient SR models. As shown in Fig.~\ref{fig:dual_connection}, we introduce a constraint on LR images to reduce the space of possible SR mapping functions. To further reduce the model redundancy,
    we propose a dual regression compression (DRC) method to compress large models (See Fig~\ref{fig:dual_compression}).
    For convenience, we term our models Dual Regression Networks (DRNs).

	\subsection{Dual Regression Learning for Super-Resolution}\label{sec:dual_scheme}
    Due to the nature of the ill-posed problems, the space of possible SR mapping functions can be extremely large, making the training very difficult.
    To alleviate this issue, we propose a dual regression learning scheme by introducing an additional constraint on LR data. 
    From Fig.~\ref{fig:dual_connection}, besides the mapping LR$\to$ HR, we also learn an inverse/dual mapping from the super-resolved images back to LR images.
    Let $\bx {\in} \mathcal{X}$ be LR images and $\by {\in} \mathcal{Y}$ be HR images.
    Unlike existing methods, we simultaneously learn a primal mapping $P$ to reconstruct HR images and a dual mapping $D$ to reconstruct LR images.
    Formally, we formulate the SR problem into the dual regression learning scheme which involves two regression tasks.
    \begin{deftn} \textbf{\emph{(Primal Regression Task for SR) }}\label{definition: primal_model}
    	We seek to find a function $P$: $\mX {\rightarrow} \mY$, such that the prediction $P(\bx)$ is similar to its corresponding HR image $\by$.
    \end{deftn}
    \begin{deftn} \textbf{\emph{({Dual} Regression Task for SR) }}\label{definition: dual_model}
    	We seek to find a function $D$: $\mY {\rightarrow} \mX$, such that the prediction of $D(\by)$ is similar to the original input LR image $\bx$.
    \end{deftn}

    \begin{figure*}[t]
    \centering
    \includegraphics[width=1.8\columnwidth]{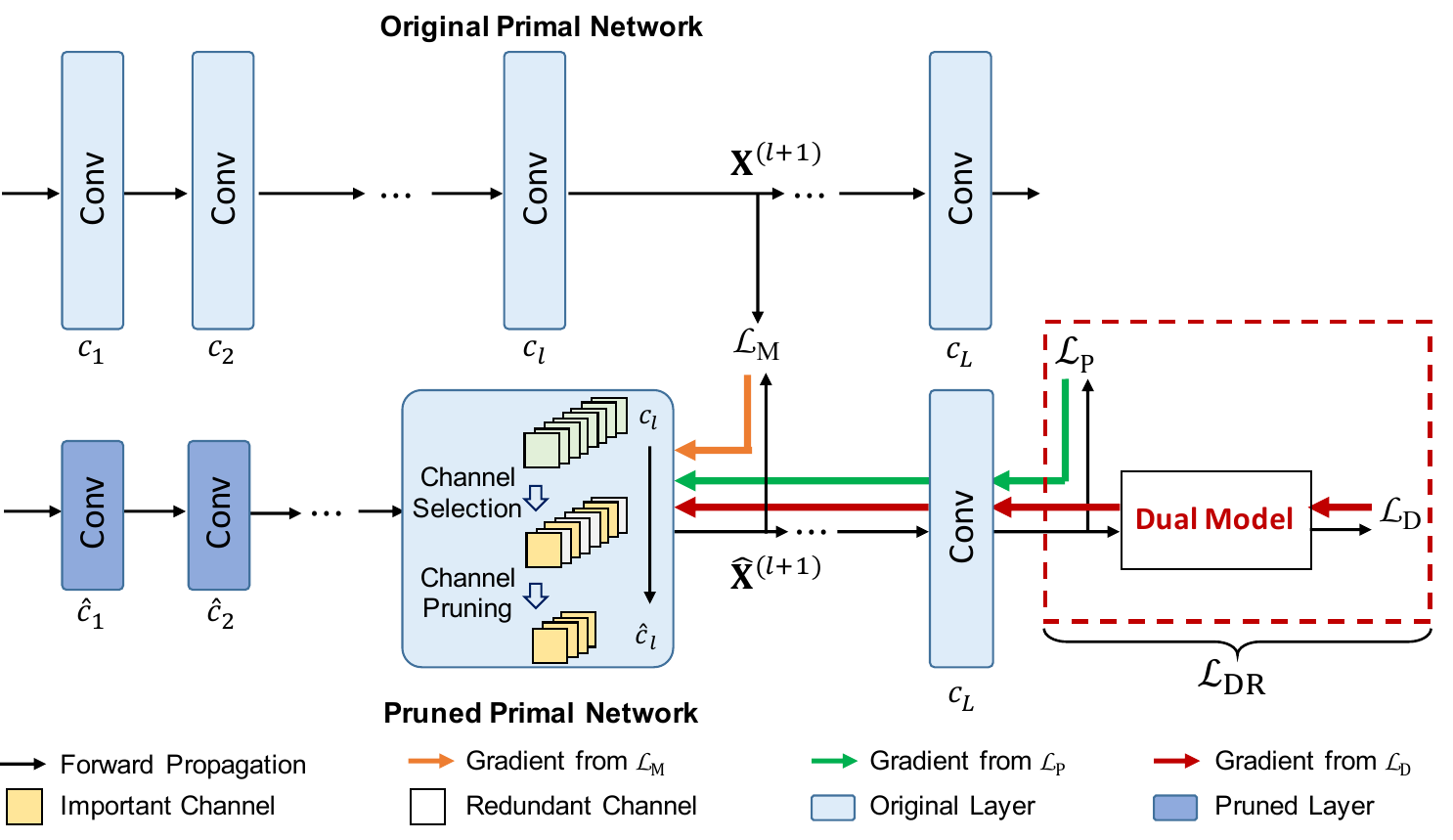}
    \caption{The dual regression based channel pruning method. We evaluate the importance of channels by computing both the feature reconstruction loss $\mL_{\rm M}$ and the dual regression loss $\mL_{\rm DR}$.
    Here, $\bX^{(l+1)}$ and $\widehat{\bX}^{(l+1)}$ denote the output features of the $l$-th layer in the original model and the pruned model, respectively. $c_l$ and $\hat{c}_l$ denote the channel number of the $l$-th layer in the original model and the pruned model.
    \dzs{The red dashed box represents our proposed dual regression loss $\mL_{\rm DR}$.}
    }
    \label{fig:dual_pruning}
    \end{figure*}

    The primal and dual learning tasks form a {closed-loop} and provide important supervision to train the models $ P $ and $ D $.
    If $P(\bx)$ was the correct HR image, then the downsampled image $D(P(\bx))$ should be very close to the input LR image $\bx$.
    By jointly learning these two tasks, we train the models based on $N$ paired samples $\left\{ (\bx_i, \by_i) \right\}_{i=1}^N$,
    where $\bx_i$ and $\by_i$ denote the $i$-th pair of LR and HR images.
    Let $\mL_{\rm P}$ and $\mL_{\rm D}$ be the loss function ($\ell_1$-norm) for the primal and dual tasks, respectively.
    The training loss becomes
    \begin{equation}
    \label{eq:dual_regression}
    \small{
    \mL_{\rm DR}(P,D) {=} \frac{1}{N}\sum_{i=1}^N \underbrace{\mL_{\rm P} \big( P(\bx_i), \by_i \big)}_{\rm primal~regression~loss}  {+} ~~\lambda \underbrace{\mL_{\rm D} \big( D(P(\bx_i)), \bx_i \big)}_{\rm dual~regression~loss}.
    }
    \end{equation}
    Here, $\lambda$ controls the weight of the dual regression loss (See the sensitivity analysis of $\lambda$ in Section~\ref{exp:lambda}).

    More critically, we also theoretically justify our method. In practice, our method has a smaller generalization bound than the vanilla training methods (\ie, without the dual mapping). In other words, our method helps to learn a more accurate LR$\to$HR mapping and improve SR performance. We summarize the theoretical analysis in Theorem~\ref{theorem:generalization_bound} and put the proof in supplementary.

    \begin{thm} \label{theorem:generalization_bound}
    	Let $\mL_{\rm DR}(P,D)$ be a mapping from $ \mX {\times} \mY $ to $ [0, 1] $ and $ \mH_{dual} $ be the function space. Let $N$ denote the number of samples and $\hat{R}_{\mZ}^{DL}$ represent the empirical Rademacher complexity\cite{mohri2012foundations} of dual learning. 
        We use $\mB(P), P{\in}\mH$ to denote the generalization bound of the supervised learning \textit{w.r.t.} the Rademacher complexity $\hat{R}_{\mZ}^{SL}(\mH)$.
    	For any error $ \delta {>} 0 $, the generalization bound of the dual regression scheme is $\mB(P, D){=}2\hat{R}_{\mZ}^{DL}(\mH_{dual}) {+} 3 \sqrt{\frac{1}{2N} \log\left(\frac{2}{\delta}\right)}$. Based on the definition of the Rademacher complexity, the capacity of the function space $ \mH_{dual}$ is smaller than the capacity of function space $ \mH $, \ie, $ \hat{R}_{\mZ}^{DL} \leq \hat{R}_{\mZ}^{SL} $.
    	In this sense, the dual regression scheme has a smaller generalization bound than the vanilla learning scheme:
    	\begin{align*}
    	    \mB(P, D) \leq \mB(P).
    	\end{align*}
    \end{thm}

    \noindent \textbf{Differences from CycleGAN based methods~\cite{zhu2017unpaired,yi2017dualgan}.} 
    Both DRN and CycleGAN~\cite{zhu2017unpaired} exploit the similar idea of building a cycle, but they have several essential differences.
    \textbf{First}, they consider different objectives. CycleGAN uses cycles to help minimize distribution divergence but DRN builds a cycle to improve reconstruction performance.
    \textbf{Second}, they consider different cycle/dual mappings. CycleGAN learns two symmetric mappings but DRN considers learning asymmetric mappings. Essentially, the primal mapping LR$\to$HR is much more complex than the dual mapping HR$\to$LR. Considering this, we design the dual model with a very small CNN (See the detailed model design in supplementary) and introduce a tradeoff parameter $\lambda$ in Eqn.~(\ref{eq:dual_regression}).
    {
    \textbf{Third}, our dual regression method is a more general scheme that can be used in more application scenarios than CycleGAN.
    Specifically, our method is a plug-and-play module that can be used to enhance diverse SR models and/or conduct model compression. 
    By contrast, CycleGAN cannot be directly applied on top of diverse SR models since it naturally requires the dual model to have the same architecture as the original/primal model. Such a large dual model may increase the training difficulty and introduce a lot of redundancy.
    Instead, our method builds a very simple dual model that is more suitable for SR tasks where the downsampling mapping is much easier to learn than the upsampling mapping. Experiments show that DRC performs well on both CNN-based and transformer-based architectures, and under both non-blind and blind SR settings (See results in Table~\ref{tab:4xsr}, ~\ref{tab:lightweight_comparison}, and~\ref{tab:compare_blind}). 
    }

\setlength{\algomargin}{0.072in}
    \begin{algorithm}[t]
        
    	\label{alg:dual_search}
    	\caption{\small Dual Regression based Channel Number Search}
    	\KwIn{Training data $\mS^{\rm train}$ and validation data $\mS^{\rm val}$;\\
    	~~~~~~~~~~~Original channel numbers $\{{c_l}\}^L_{l=1}$;\\
    	~~~~~~~~~~~Channel number configurations $\{\alpha_l\}_{l=1}^{L}$;\\
    	~~~~~~~~~~~Candidate scaling factors $\mV$;\\
    	~~~~~~~~~~~Target compression ratio $r$.}
    	\KwOut{Searched channel numbers $\{\hat{c}_l\}^L_{l=1}$.}
        Rebuild model $\boldsymbol{\alpha}$ with configuration parameters $\{\alpha_l\}_{l=1}^{L}$;\\
    	\While{$not ~ converge$}{
    	    // \emph{Update the channel number configuration $\boldsymbol{{\alpha}}$} \\
        	Sample data batch from $\mS^{\rm val}$ to compute {$\mL^{\rm val}_{\rm DR}$};\\
            Evaluate channel importance and rank the channels; \\
        	Update $\boldsymbol{{\alpha}}$ by descending {$\nabla_{{\alpha}} ~\mL^{\rm val}_{\rm DR} \big( (\boldsymbol{{\alpha}}; \bW^*), D \big)$}; \\
    	    // \emph{Update model parameters $\bW$} \\
    	    Sample data batch from $\mS^{\rm train}$ to compute {$\mL^{\rm train}_{\rm DR}$};\\
        	Update $\bW$ by descending {$\nabla_{\bW} ~\mL^{\rm train}_{\rm DR} \big( (\boldsymbol{{\alpha}}; \bW), D \big)$};\\
        }
        \For{$l = 1$ {\rm to} $L$}{
        Select the scaling factor $\hat{v} = \argmax_{v \in \mV} \alpha_l^{(v)}$;\\
        Compute the channel number $\hat{c}_l = c_l \cdot (1 - r) \cdot \hat{v}$;
        }
    \end{algorithm}

	\subsection{Dual Regression Compression}\label{sec:dual_compression}
    Most SR models have extremely high computational cost and cannot be directly deployed to the devices with limited computation resources.  
    To alleviate this issue, one can apply model compression techniques to obtain lightweight models. However, it is non-trivial to accurately identify the redundant components (\eg, layers or channels) due to the extremely large mapping space. Specifically, once we learn an inaccurate SR mapping, the predicted model redundancy may be also inaccurate, leading to significant performance drop (See results in supplementary). 

	To address the above issues, we build a lightweight dual regression compression method based on channel pruning techniques to compress SR models in a reduced mapping space.
    Let $\psi(\cdot)$ be the function to measure the computational cost of models (\eg, the number of parameters).
	Given a primal model $P$ and a target compression ratio $r$, we seek to obtain a compressed model $\widehat{P}$ that satisfies the constraint $\psi(\widehat{P}) \le (1-r) \psi(P)$. 
	Supposing that both $P$ and $\widehat{P}$ share the same dual model $D$,
    the optimization problem becomes:
	\begin{equation} \label{eq:framework_optimization}
	\min_{\widehat{P}} ~ \mL_{\rm DR}(\widehat{P}, D) ~~~{\rm s.t}.~~ \psi(\widehat{P}) \le (1-r) \psi(P).
	\end{equation}
	
    In this paper, we seek to reduce the model redundancy in both layer-level and channel-level.
    As shown in Fig.~\ref{fig:dual_compression}, we first determine the redundancy for each layer by performing dual regression channel number search. Then, we exploit the dual regression scheme to evaluate the importance of channels and prune the redundant ones according to the searched channel numbers.

    \begin{figure*}[htpb]
        \centering
        \subfigure[Independent channel number search strategy.]{
        \label{fig:independent_search}
        \includegraphics[width=0.45\textwidth]{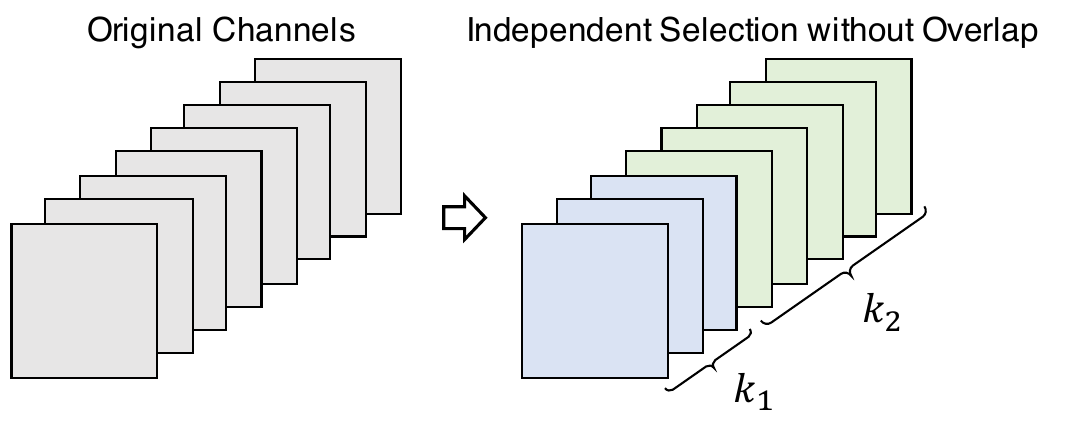}
        }
        \subfigure[Channel-wise weight-sharing strategy.]{
        \label{fig:weight_sharing}
        \includegraphics[width=0.45\textwidth]{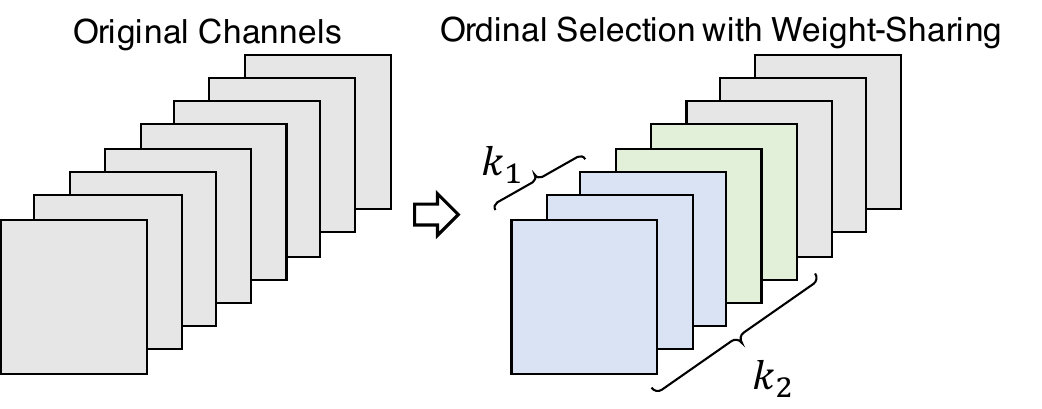}
        }
        \subfigure[Overview of our importance-aware search strategy.]{
        \label{fig:importance_aware}
        \includegraphics[width=0.97\textwidth]{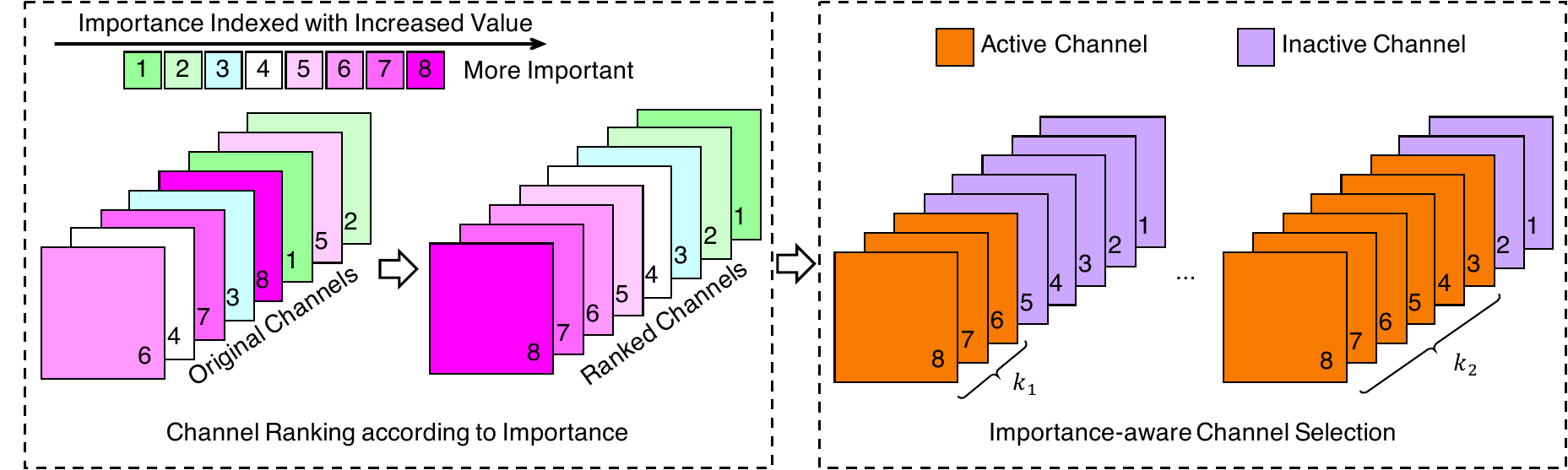}
        }
        \caption{
        Overview of most channel number search strategies and our importance-aware search strategy. 
        (a) Some previous works~\cite{peng2019efficient, fang2020densely} assume that different channel configurations should be treated individually. For two candidate numbers of channels $k_1$ and $k_2$ ($k_1 < k_2$), the selected $k_2$ channels are independent of the $k_1$ channels.
        (b) Some other works~\cite{guo2020single, wan2020fbnetv2, wang2020revisiting} use the weight-sharing strategy to reduce the search cost. For two candidate numbers of channels $k_1$ and $k_2$ ($k_1 < k_2$), the selected $k_2$ channels contain all the $k_1$ channels. The weights of these overlapped channels are shared across different sets of channels during searching.
        (c) Based on the weight-sharing strategy, we further propose an importance-aware search strategy to search for a promising/suitable channel configuration to recognize and reduce layer-wise redundancy. For each candidate number of channels $k$, we select top-$k$ important channels and ignore the rest of the redundant channels simultaneously.
        \dzs{Note that we keep the position of the selected $k$ channels on the original model unchanged, avoiding the effect of the ranking operation on the output features.}
        }
    \label{fig:importance_search}
    \end{figure*}

    \subsubsection{Dual Regression based Channel Number Search}
    
    Most channel pruning methods adopt a hand-crafted compression policy to prune deep models~\cite{li2017pruning, he2018soft},
    \eg, pruning 50\% channels in all the layers. 
    However, such a compression policy may not be optimal since different layers often have different redundancy~\cite{he2018amc}.
    To address this issue, we propose a dual regression channel number search method to recognize the redundancy of each layer by determining the promising number of channels to be preserved.
    {To save the training cost, we follow~\cite{wang2020revisiting,wan2020fbnetv2} and use the weight-sharing strategy. Unlike existing search methods, we propose an importance-aware search strategy to select channels according to their importance during the search process. 
    }
    We show the details of the proposed method in Algorithm~\ref{alg:dual_search}.

    Given a primal model $P$ with $L$ layers, we use $\{c_l\}_{l=1}^{L}$ to denote the channel numbers of different layers. 
    To obtain a model that satisfies the target compression ratio $r$, for any layer $l$, we first remove $c_l \cdot r$ channels 
    and then investigate whether we can further remove more channels without degrading the performance. 
    Nevertheless, the search space would be extremely large since the candidate channel number can be any positive integer lower than $c_l$.
    To alleviate this issue,
    we construct the search space by considering a set of candidate scaling factors $\mV = \{ 50\%, 60\%, 70\%, 80\%, 90\%, 100\% \}$ to scale the channel number. 
    Specifically, for the $l$-th layer, we seek to select a scaling factor $\hat{v} \in \mV$ to obtain the resultant channel number $\hat{c}_l = c_l \cdot (1 - r) \cdot \hat{v}$ in the compressed model.
    {
    Notably, for the $l$-th layer, we use $r$ to define the maximum number of channels $c_l \cdot (1 - r)$, which acts as the upper bound of the candidate channel numbers during the search process.
    We use six $v$ for each layer to define the search space of candidate channel numbers, which enables selecting different channel numbers that are not larger than the upper bound.
    }

    {
    Given a target compression ratio $r$ and a candidate scaling factor $v$, we need to select $k = c_l \cdot (1 - r) \cdot v$ channels to build the $l$-th layer. 
    As shown in Fig.~\ref{fig:importance_search}\subref{fig:independent_search} and ~\ref{fig:importance_search}\subref{fig:weight_sharing}, most search methods do not consider the importance of channels when determining the optimal channel configuration. The ignoration of channel importance may cause inconsistency between channel number search and channel pruning, resulting in limited performance (See Table~\ref{tab:importance_aware}).
    To address this issue, we adopt the consistent criterion of evaluating channel importance during both the search process and the following pruning process.
    As shown in Fig.~\ref{fig:importance_search}\subref{fig:importance_aware}, we first rank the channels according to their importance in terms of the $\ell_1$-norm on the weights of each channel. 
    Then, we selected the top-$k$ important channels based on the ranking result. 
    In this sense, we optimize the top-$k$ important channels instead of only the first $k$ channels. 
    We highlight that the importance rank of channels can be continuously updated during the search process since we also optimize the weights at the same time. 
    }

    To find the promising channel {configurations}, we adopt the differentiable search strategy~\cite{liu2018darts} by relaxing the search space to be continuous.
    For any layer $l$, we construct a channel number configuration $\alpha_l \in \mathbb{R}^{|\mV|}$ in which each element $\alpha_l^{(v)}$ indicates the importance of a specific scaling factor $v$.
    For any layer $l$, let $\bX^{(l)}$ be the input features, $\bW^{(l)}$ be the parameters, and $\otimes$ be the convolutional operation.
    For convenience, we use $\bX_{[1:k]}^{(l)}$ and $\bW_{[1:k]}^{(l)}$ to denote the features and parameters \wrt 
    {the top-$k$ important channels (see the channel selection in Fig.~\ref{fig:importance_search}\subref{fig:importance_aware}).}
    In this paper, we use ${c}^{(v)}_l$ to denote the number of channels specified by a specific scaling factor $v$.
    Following~\cite{liu2018darts}, we relax the categorical choice of a particular factor as a softmax over all possible factors. 
    {
    Then, we use the sum operation to make the search space continuous and the search process differentiable.
    Formally, the output of the $l$-th layer is
    }
    \begin{equation}
    \bX^{(l+1)} = \sum_{v \in \mV} \frac{\exp(\alpha^{(v)}_l)}{\sum_{v' \in \mV} \exp(\alpha^{(v')}_l)}   \bX^{(l)}_{[1: {c}^{(v)}_l]} \otimes \bW^{(l)}_{[1: {c}^{(v)}_l]}~~.
    \end{equation}
    With the continuous relaxation, the task of channel number search becomes learning a set of continuous variables $\boldsymbol{\alpha} = \{\alpha_l\}^L_{l=1}$.
    As shown in Algorithm~\ref{alg:dual_search}, for any layer $l$, we obtain the resultant channel numbers by selecting the most likely element in $\alpha_l$.

    {
    Due to the extremely large function space of the ill-posed SR problem that contains a lot of undesired blurry solutions, it is non-trivial to accurately identify the redundancy of each layer for most search methods.}
    To enhance the search process, we minimize the dual regression loss to reduce the space of possible mapping functions.
    Let $\mL^{\rm train}_{\rm DR}$ and $\mL^{\rm val}_{\rm DR}$ be the dual regression loss computed on the training data and validation data.
    Here, we use the continuous variables $\boldsymbol{\alpha}$ and the model parameters $\bW$ to represent the primal model $P=(\boldsymbol{\alpha};\bW)$.
    Given a dual model $D$, the optimization problem of channel number search becomes
    \begin{equation} \label{eq:bilevel_optimization}
    \begin{aligned}
    & \min_{\boldsymbol{{\alpha}}} ~ \mL^{\rm val}_{\rm DR} \big( (\boldsymbol{{\alpha}}; \bW^*), D \big) \\
    & {\rm s.t.} ~~ \bW^* = \argmin_{\bW} ~\mL^{\rm train}_{\rm DR} \big( (\boldsymbol{{\alpha}}; \bW), D \big).
    \end{aligned}
    \end{equation}

    \noindent \textbf{{Differences from DARTS~\cite{liu2018darts}}}: 
    {Our DRC has essential differences from DARTS.
    \textbf{First}, we design an importance-aware search strategy to facilitate the channel number search for pruning. Instead, DARTS are designed for architecture design, which does not involve important channel selection. Therefore, it is easier for our proposed DRS than DARTS to identify the channel redundancy for each layer. Our searched model outperforms that searched by DARTS significantly (see results in Table~\ref{tab:importance_aware}). 
    \textbf{Second}, the formulation of our dual regression-based channel number search (DRS) is different from DARTS. 
    Our DRS is built upon our dual regression scheme, which helps to constrain the complex possible function space of the ill-posed SR optimization problem (see Theorem~\ref{theorem:generalization_bound}).
    Thus, our DRS makes it easier to identify the layer-wise redundancy for channel pruning, resulting in lightweight SR models without significant performance degradation after channel pruning. 
    Instead, without using the dual regression scheme, DARTS makes it hard to recognize the redundancy of SR models accurately. Thus, it may only obtain less compact models with larger performance degradation  (see results in Table~\ref{tab:importance_aware}).
    }

    \begin{table*}
    \centering
    \caption{Performance comparisons for 4$\times$ image super-resolution.
    ``-'' denotes the results that are not reported. We highlight that our dual regression (DR) method is able to enhance both CNN-based and transformer-based SR models, showing a high flexibility on top of diverse architectures.}
    \label{tab:4xsr}
    \resizebox{0.99\textwidth}{!}
    {
    \begin{tabular}{c|c|c|c|c|c|c|c} 
    \toprule
    \multirow{2}{*}{Method}  &   \multirow{2}{*}{\#Params (M)} & \multirow{2}{*}{\#Madds (G)}  &  Set5                            & Set14                           & BSDS100                        & Urban100               & Manga109                         \\
    &     &                       &                    PSNR / SSIM                     & PSNR / SSIM                     & PSNR / SSIM                     & PSNR / SSIM            & PSNR / SSIM                      \\ 
    \hline
    Bicubic      &  - & -                                                            & 28.42 / 0.810 & 26.10 / 0.702 & 25.96 / 0.667 & 23.15 / 0.657 & 24.92 / 0.789  \\
    ESPCN~\cite{shi2016real}         & - &  0.2                       & 29.21 / 0.851 & 26.40 / 0.744 & 25.50 / 0.696 & 24.02 / 0.726 & 23.55 / 0.795  \\
    LapSRN~\cite{lai2017deep}        & 0.9 & 29.2                           & 31.54 / 0.885 & 28.09 / 0.770 & 27.31 / 0.727 & 25.21 / 0.756 & 29.09 / 0.890  \\
    DRRN~\cite{tai2017image}      & 0.3 & 1087.5                                   & 31.68 / 0.889 & 28.21 / 0.772 & 27.38 / 0.728 & 25.44 / 0.764 & 29.46 / 0.896  \\
    CARN~\cite{ahn2018fast}      & 1.1 & 14.6                              & 32.13 / 0.894 & 28.60 / 0.781 & 27.58 / 0.735 & 26.07 / 0.784 & 30.47 / 0.908  \\
    IMDN~\cite{hui2019lightweight}      & 0.7 & 6.6                       & 32.21 / 0.895 & 28.58 / 0.781 & 27.56 / 0.735 & 26.04 / 0.784 & 30.45 / 0.908  \\
    PAN~\cite{zhao2020efficient}    & 0.3 &  4.5                         & 32.13 / 0.895 & 28.61 / 0.782 & 27.59 / 0.736 & 26.11 / 0.785 & 30.51 / 0.910  \\
    SRResNet~\cite{ledig2017photo}        & 1.5 &   20.5                     & 32.05 / 0.891 & 28.49 / 0.782 & 27.61 / 0.736 & 26.09 / 0.783 & 30.70 / 0.908  \\
    SRGAN~\cite{ledig2017photo}      & 1.5 &   20.5                       & 29.46 / 0.838 & 26.60 / 0.718 & 25.74 / 0.666 & 24.50 / 0.736 & 27.79 / 0.856  \\
    ECBSR~\cite{zhang2021edge}        & 0.6 &  5.5                       & 31.92 / 0.895 & 28.34 / 0.782 & 27.48 / 0.739 & 25.81 / 0.777 & -              \\
    SMSR~\cite{wang2021exploring}        & 1.0 &   -                   & 32.12 / 0.893  & 28.55 / 0.781  & 27.55 / 0.735  & 26.11 / 0.787  & 30.54 / 0.909   \\
    SR-APS~\cite{zhan2021achieving}     & 0.1 &   -                      & 31.93 / 0.891  & 28.42 / 0.776  & 27.44 / 0.731   & 25.66 / 0.772   & -              \\
    DFSR~\cite{zhang2021data}     & 10.8 &    116.1                        & 31.78 / 0.890  & 28.33 / 0.7758  & 27.38 / 0.729   & 25.40 / 0.761  & -              \\
    SRDenseNet~\cite{tong2017image}       & 2.0    & 62.3                   & 32.02 / 0.893 & 28.50 / 0.778 & 27.53 / 0.733 & 26.05 / 0.781 & 29.49 / 0.899  \\
    EDSR~\cite{lim2017enhanced}      & 43.1      & 463.1                         & 32.48 / 0.898 & 28.81 / 0.787 & 27.72 / 0.742 & 26.64 / 0.803 & 31.03 / 0.915  \\
    DBPN~\cite{DBPN2018}         & 15.3     & 1220.4                               & 32.42 / 0.897 & 28.75 / 0.786 & 27.67 / 0.739 & 26.38 / 0.794 & 30.90 / 0.913  \\
    RCAN~\cite{zhang2018image}    & 15.6  & 147.1                            & 32.63 / 0.900 & 28.85 / 0.788 & 27.74 / 0.743 & 26.74 / 0.806 & 31.19 / 0.917  \\
    RRDB~\cite{wang2018esrgan}     & 16.7      & 165.2                 & 32.73 / 0.901 & 28.97 / 0.790 & 27.83 / 0.745 & 27.02 / 0.815 & 31.64 / 0.919  \\
    HAN~\cite{niu2020single}        & 16.2    & 151.5                        & 32.61 / 0.900 & 28.90 / 0.789 & 27.79 / 0.744 & 26.85 / 0.809 & 31.44 / 0.918  \\
    CSNLN~\cite{mei2020image}      & 6.6   & 4428.5                             & 32.68 / 0.900 & 28.95 / 0.789 & 27.80 / 0.744 & 27.22 / 0.817 & 31.43 /0.920   \\
    SAN~\cite{dai2019second}    & 15.8    & 150.1                             & 32.64 / 0.900 & 28.92 / 0.788 & 27.79 / 0.743 & 26.79 / 0.806 & 31.18 / 0.916  \\
    ClassSR~\cite{kong2021classsr}    & 30.1 &        -         & 32.25 / 0.898  & 28.77 / 0.788  & 27.65 / 0.741  & 26.70 / 0.804  & 31.17 / 0.916   \\
    SwinIR~\cite{liang2021swinir}    & 11.9 &  121.1            & 32.92 / 0.904  & 29.09 / 0.795  & 27.92 / 0.749  & 27.45 / 0.825  & 32.05 / 0.926   \\
    DAT~\cite{chen2023dual}        &  14.8 &  155.1       & 33.08 / 0.906   & 29.23 / 0.797   & 28.00 / 0.751  & 27.87 / 0.834 &  32.51 / 0.929   \\
    \hline
    DRN-S    & 4.8   & 109.9 & 32.68 / 0.901 & 28.93 / 0.790 & 27.78 / 0.744 & 26.84 / 0.807 & 31.52 / 0.919  \\ 
    DRN-L     & 9.8   & 224.8  & 32.74 / 0.902 & 28.98 / 0.792 & 27.83 / 0.745 & 27.03 / 0.813 & 31.73 / 0.922  \\
    SwinIR-DR  & 11.9 &  121.1 & 33.03 / 0.904 &	29.19 / 0.795 & 27.98 / 0.747 & 27.80 / 0.831 &	32.38 / 0.925  \\
    DAT-DR  &   14.8 &  155.1 &   33.17 / 0.906	& 29.30 / 0.798	& 28.04 / 0.752 &	28.04 / 0.837 &	32.71 / 0.930   \\

    \bottomrule
    \end{tabular}}
    \end{table*}
    
    \subsubsection{Dual Regression based Channel Pruning}\label{sec:dual_pruning}
    Based on the searched channel numbers, we still need to determine which channels should be pruned. 
    One of the key challenges is how to accurately evaluate the importance of channels.
    To address this, we develop a dual regression channel pruning method that exploits the dual regression scheme to identify the important channels.
    We show our method in Fig.~\ref{fig:dual_pruning}. 

    Let $P$ and $\widehat{P}$ be the original primal model and the compressed model, respectively.
    We use $\bX^{(l+1)}$ and $\widehat{\bX}^{(l+1)}$ to denote the output feature maps of the $l$-th layer in $P$ and $\widehat{P}$. 
    Given the searched channel numbers $\{\hat{c}_l\}_{l=1}^L$, 
    we seek to select the channels which really contribute to SR performance.
    Nevertheless, this goal is non-trivial to achieve due to the extremely large mapping space incurred by the ill-posed problem.
    To address this issue,
    we exploit the dual regression scheme to evaluate the importance of channels. 
    Specifically, we consider one channel as an important one if it helps to reduce the dual regression loss $\mL_{\rm DR}(\widehat{P}, D)$. 
    Moreover, for any layer $l$, we also minimize the reconstruction error~\cite{luo2017thinet,li2017pruning} of the feature maps between $P$ and $\widehat{P}$, \ie, $\mL_{\rm M}(\bX^{(l+1)}, \widehat{\bX}^{(l+1)})$, to further improve the performance.
    Given a specific channel number $\hat{c}_l$, we impose an $\ell_{0}$-norm constraint $\| \bW^{(l)} \|_{0} {\leq} \hat{c}_l$ on the number of active channels in $\bW^{(l)}$.
    Formally, the channel pruning problem for the $l$-th layer is:
    \begin{equation}\label{eq:pruning_optimization}
        \min_{\bW^{(l)}} ~\mL_{\rm M}(\bX^{(l+1)}, \widehat{\bX}^{(l+1)}) + \gamma \mL_{\rm DR}(\widehat{P}, D), ~~~ {\rm s.t.} ~ \| \bW^{(l)} \|_{0} \leq \hat{c}_l,
    \end{equation}
    where $\gamma$ is a hyper-parameter that controls the weight of the dual regression loss (See more discussions on $\gamma$ in Section~\ref{exp:gamma}).
    However, Problem~(\ref{eq:pruning_optimization}) is hard to solve due to the training difficulty incurred by the $\ell_{0}$-norm constraint. To address this, we adopt a greedy strategy~\cite{zhuang2018nips,bahmani2013greedy,liu2021discrimination} in which we first remove all the channels and then select the most important channels one by one.
    Following~\cite{zhuang2018nips}, we perform channel selection according to the gradients \wrt different channels. 

    \begin{figure*}[t!] 
        \centering
    		\includegraphics[width = 1.0\textwidth]{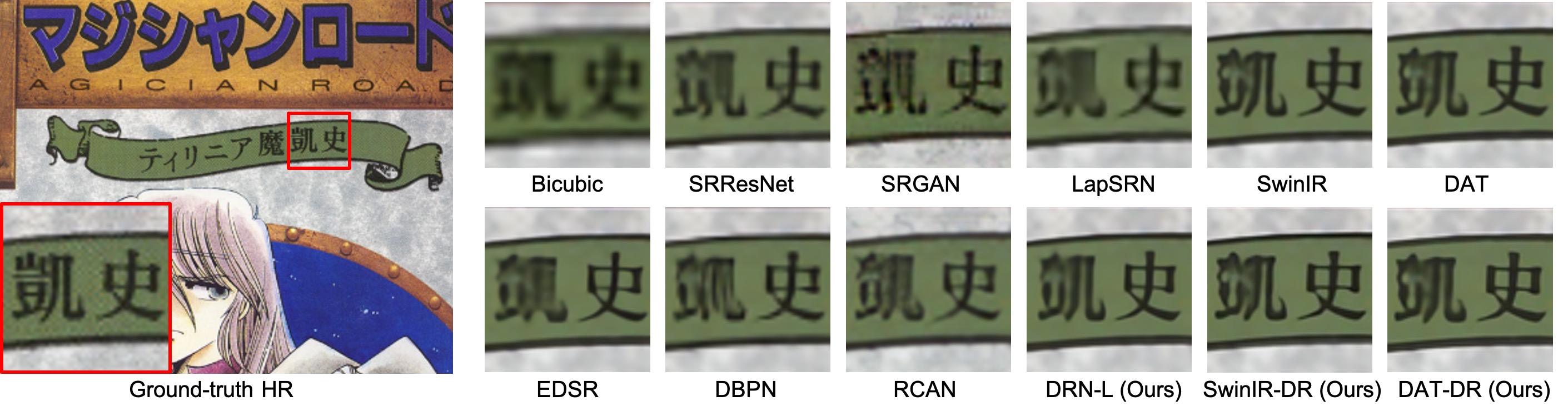}\label{fig:img4x_compare1_sup} 
    	\caption{Visual comparisons of the images produced by different models for $4\times$ image super-resolution on benchmark datasets. We show that all the models enhanced by our DR consistently produce sharper images, i.e., with more high-frequency information, than their original counterparts.}
    	\label{fig:image_compare_4x}
    \end{figure*}

    \section{Experiments}\label{sec:exp}
    We conduct extensive experiments to verify the effectiveness of the proposed methods.
    First, we evaluate the proposed dual regression (DR) learning scheme.
    Second, we compare the proposed dual regression compression (DRC) method with different model compression methods. Third, we further apply our DRC method to compress blind SR models.
    The source code is available at \href{https://github.com/guoyongcs/DRC}{https://github.com/guoyongcs/DRC}.


    \subsection{Datasets and Implementation Details}
    Based on the dual regression scheme, we build our DRN based on the design of U-Net for SR~\cite{iqbal2019super,ronneberger2015u} (See more details in the supplementary materials).  
    We first propose two models, including a small model DRN-S and a large model DRN-L. 
    {
    We also apply our dual regression scheme to two popular SR models, i.e., SwinIR~\cite{liang2021swinir} and DAT~\cite{chen2023dual}.
    Then, we use the proposed dual regression compression method to compress the DRN-S and SwinIR-light~\cite{liang2021swinir} models. By default, we consider the target pruning ratio of 30\% in most experiments. We also conduct additional experiments with other pruning ratios $r \in \{30\%, 50\%, 70\%\}$ and put the results in supplementary.
    }

    \textbf{Datasets and evaluation metrics.} 
    Following~\cite{wang2018esrgan}, we train our models on DIV2K~\cite{timofte2017ntire} and Flickr2K~\cite{lim2017enhanced} datasets, which contain $800$ and $2650$ images separately.
    For quantitative comparison, we evaluate different SR methods on five benchmark datasets, including Set5~\cite{bevilacqua2012low}, Set14~\cite{zeyde2010single}, BSDS100~\cite{arbelaez2011contour}, Urban100~\cite{huang2015single} and Manga109~\cite{matsui2017sketch}. 
    {Moreover, for the blind SR setting where the degradation model of each test LR image is unknown, we use the DIV2KRK~\cite{bell2019blind} dataset to evaluate the performance of different blind SR methods.}
    \qi{To assess the quality of super-resolved images, we adopt two commonly used metrics, \ie, {PSNR} and {SSIM}~\cite{wang2004image}.}
    {The computational cost \#MAdds are measured on a 96 $\times$ 96 LR image.}
    
    \textbf{Training details.} 
    During the training of our DRN-S and DRN-L, we apply Adam~\cite{kingma2014adam} with $\beta_1 = 0.9$, $\beta_2 = 0.99$ and set minibatch size as 32.
    We use RGB input patches with size $48 \times 48$ from LR images and the corresponding HR patches as the training data, and augment the training data following the method in~\cite{lim2017enhanced, zhang2018image}.
    The learning rate is initialized to $10^{-4}$ and decreased to $10^{-7}$ with a cosine annealing strategy.
    {
    To train the SwinIR-DR and DAT-DR models, we follow the recipe of the training setting of SwinIR~\cite{liang2021swinir} and DAT~\cite{chen2023dual}, respectively.
    As for model compression, following the pruning pipeline of~\cite{zhuang2018nips,liu2021discrimination}, we take two lightweight models DRN-S and SwinIR-light~\cite{liang2021swinir} as the baselines, and conduct pruning followed by finetuning.
    For the blind SR setting, we randomly generate anisotropic Gaussian kernels to synthetic LR images from the given HR images following the setting in DCLS~\cite{luo2022deep}. 
    We use the synthetic LR-HR paired data to train and compress the baseline DCLS~\cite{luo2022deep} model to obtain the lightweight DCLS-DRC model.
    }

\begin{table*}[htbp]
  \centering
  \caption{
  {Comparison with lightweight SR models and pruning methods on five benchmarks for 4$\times$ SR. We consider the pruning ratio of 30\% for all the pruning methods for fair comparison. We highlight that our pruned model consistently outperforms all the considered lightweight SR models as well as the pruning methods.}
  }
    \begin{tabular}{c|c|c|c|c|c|c|c|c}
    \toprule
    \multicolumn{1}{c|}{\multirow{2}[0]{*}{Model}} & \multicolumn{1}{c|}{\multirow{2}[0]{*}{Method}} & \multicolumn{1}{c|}{\multirow{2}[0]{*}{\#Params (K)}} & \multicolumn{1}{c|}{\multirow{2}[0]{*}{\#Madds (G)}} & \multicolumn{1}{c|}{Set5} & \multicolumn{1}{c|}{Set14} & \multicolumn{1}{c|}{BSDS100} & \multicolumn{1}{c|}{Urban100} & \multicolumn{1}{c}{Manga109} \\
          &       &       &       & \multicolumn{1}{c|}{PSNR / SSIM} & \multicolumn{1}{c|}{PSNR / SSIM} & \multicolumn{1}{c|}{PSNR / SSIM} & \multicolumn{1}{c|}{PSNR / SSIM} & \multicolumn{1}{c}{PSNR / SSIM} \\
    \hline
    
    \multicolumn{2}{c|}{SRCNN~\cite{dong2016image}}          & 57                           & 8.5                                                       & 30.48 / 0.863 & 27.49 / 0.750 & 26.90 / 0.710 & 24.52 / 0.722 & 27.66 / 0.851  \\
    \multicolumn{2}{c|}{FSRCNN~\cite{dong2016accelerating}}  & 12                           & 0.7                                & 30.71 / 0.866 & 27.59 / 0.754 & 26.98 / 0.710 & 24.62 / 0.728 & 27.90 / 0.852  \\
    \multicolumn{2}{c|}{CARN-M~\cite{ahn2018fast}}           & 300                           & 5.2                             & 31.92 / 0.890 & 28.42 / 0.776 & 27.44 / 0.730 & 25.62 / 0.769 & 25.62 / 0.769  \\
    \multicolumn{2}{c|}{VDSR~\cite{kim2016accurate}} & 665 & 612.6 & 31.35 / 0.883 & 28.01 / 0.767 & 27.29 / 0.725 & 25.18 / 0.752 & - \\
    \multicolumn{2}{c|}{LapSRN~\cite{lai2017deep}} & 813 & 149.4 & 31.54 / 0.885 & 28.19 / 0.772 & 27.32 / 0.728 & 25.21 / 0.756 & - \\
    \multicolumn{2}{c|}{DRRN~\cite{tai2017image}} & 297 & 6,796.9 & 31.68 / 0.888 & 28.21 / 0.772 & 27.38 / 0.728 & 25.44 / 0.763 & - \\
    \multicolumn{2}{c|}{MemNet~\cite{tai2017memnet}} & 677 & 2,662.4 & 31.74 / 0.889 & 28.26 / 0.772 & 27.40 / 0.728 & 25.50 / 0.763 & - \\
    \multicolumn{2}{c|}{IMDN~\cite{hui2019lightweight}} & 715 & 40.9 & 32.21 / 0.894 & 28.58 / 0.781 & 27.56 / 0.735 & 26.04 / 0.783 & 30.45 / 0.907  \\
    \multicolumn{2}{c|}{LAPAR-A~\cite{li2020lapar}} & 659 & 94.0 & 32.15 / 0.894 & 28.61 / 0.781 & 27.61 / 0.736 & 26.14 / 0.787 & 30.42 / 0.907 \\
    \multicolumn{2}{c|}{LatticeNet~\cite{luo2020latticenet}} & 777 & 43.6 & 32.30 / 0.896 & 28.68 / 0.783 & 27.62 / 0.736 & 26.25 / 0.787 & - \\
    \multicolumn{2}{c|}{DI-trim-0.3-SRResNet~\cite{hou2020efficient}} &  604  & 14.9 & 32.32 / 0.894   &  28.70 / 0.783     &   27.60 / 0.737    &    26.14 / 0.786 &  -   \\      
    \multicolumn{2}{c|}{SMSR~\cite{wang2021exploring}} &  1006     &   -  &   32.12 / 0.893    &   28.55 / 0.781    &    27.55 / 0.735   &    26.11 / 0.787   &  30.54 / 0.909 \\      
    \multicolumn{2}{c|}{NAPS~\cite{zhan2021achieving}} &  125     &   7.1    &   31.93 / 0.867    &   27.68 / 0.756    &    26.98 / 0.716   &   24.65 / 0.730    & - \\      
    \multicolumn{2}{c|}{SRPN-Lite~\cite{zhang2022learning}} &  623     &   35.8    &   32.24 / 0.896    &   28.69 / 0.784    &    27.63 / 0.737   &   26.16 / 0.788    & - \\   

    \multicolumn{2}{c|}{ELAN-light~\cite{zhang2022efficient}} & 640 & 53.72 & 32.43 / 0.897 & 28.78 / 0.785 & 27.69 / 0.740 & 26.54 / 0.798 & 30.92 / 0.915 \\
    \hline
        \multirow{6}[0]{*}{DRN-S} & Baseline    & 4800  & 109.9  & {32.68} / {0.901}  & {28.93} / {0.790}  & {{27.78}} / {{0.744}}  & {{26.84}} / {{0.807}}  & {{31.52}} / {{0.919}}  \\
        \cline{2-9}
              & CP~\cite{he2017channel}    &   3473    &   77.4    &   32.26 / 0.887    &   28.47 / 0.777    &    27.38 / 0.732   &   26.42 / 0.794    & 31.03 / 0.904 \\
              & ThiNet~\cite{Luo2019ThiNet}  &   3473    &   77.4 &     32.33 / 0.889  &   28.57 / 0.781    &    27.45 / 0.734   &   26.53 / 0.796    &  31.12 / 0.907 \\
              & DCP~\cite{zhuang2018nips} &   3473    &   77.4 &  32.41 / 0.893     &   28.65 / 0.782    &   27.52 / 0.737    &    26.61 / 0.800   & 31.21 / 0.912 \\
              & SRP~\cite{zhang2022learning} & 3473    &   77.4 &  32.49 / 0.894 & 28.77 / 0.786 & 27.63 / 0.740 & 26.80 / 0.806 & 31.44 / 0.917  \\
              & DRC (Ours) &   3116    & 72.3      &   {\textbf{32.66}} / {\textbf{0.900}}    &   {\textbf{28.92}} / {\textbf{0.789}}    &  \textbf{27.82} / \textbf{0.744}     &   \textbf{26.95} / \textbf{0.809}    &  \textbf{31.64} / \textbf{0.921} \\
    \hline
        \multirow{6}[0]{*}{SwinIR-light~\cite{liang2021swinir}} & Baseline  &  897 & 10.0 &  32.44 / 0.897 & 28.77 / 0.785 & 27.69 / 0.740 & 26.47 / 0.798 & 30.92 / 0.915  \\
        \cline{2-9}
              & CP~\cite{he2017channel}    &    689   &   7.3    &  32.05 / 0.883     &   28.34 / 0.771   & 27.36 / 0.730      &   26.13 / 0.789    &  30.57 / 0.910  \\
              & ThiNet~\cite{Luo2019ThiNet} &   689   &   7.3   &    32.24 / 0.888   &   28.57 / 0.775    &  27.42 /  0.732    &   26.25 / 0.791    & 30.74 / 0.912 \\
              & DCP~\cite{zhuang2018nips}   &   689   &   7.3     &   32.27 /  0.891   &   28.66 / 0.780    &   27.52 / 0.735    &  26.31 / 0.793     &  30.79 / 0.911 \\
              & SRP~\cite{zhang2022learning} & 689   &   7.3 & 30.31 / 0.892 &  28.71 / 0.781 & 27.51 / 0.735 & 26.33 / 0.794 & 30.82 / 0.913  \\
              & DRC (Ours) &    635   &   6.8    & \textbf{32.44} / \textbf{0.896}      &  \textbf{28.81} / \textbf{0.785}     &   \textbf{27.70} / \textbf{0.738}    &   \textbf{26.44} / \textbf{0.797}    &  \textbf{30.94} / \textbf{0.915} \\
    \bottomrule
    \end{tabular}%
  \label{tab:lightweight_comparison}%
\end{table*}%

    \textbf{Details of channel number search and channel pruning.} 
    \qi{We search the channel numbers in each layer for the compressed models on DIV2K~\cite{timofte2017ntire} dataset.}
    {We primarily consider removing $r = 30\%$ to obtain more lightweight SR models for both our DRN-S and the SwinIR-light~\cite{liang2021swinir} models.}
    Following~\cite{liu2018darts}, we use zero initialization for the continuous variables $\boldsymbol{{\alpha}}$, which ensures $\boldsymbol{{\alpha}}$ to receive sufficient learning signal at the early stage.
    We use Adam~\cite{kingma2014adam} optimizer to train the model with the learning rate $\eta = 3\times10^{-4}$ and the momentum $\beta=(0.9, 0.999)$.
    We train the channel number search model for 100 epochs with a batch size of $16$.
    {The channel number search process takes approximately 10 hours on a TITAN A100 GPU.}
    \qi{As for channel pruning, we perform dual regression channel pruning to select important channels on DIV2K~\cite{timofte2017ntire} dataset.}
    During pruning, once we remove the input channels of the $l$-th convolution layer, the output channels of the previous convolution layer can be removed correspondingly. 
    Once a new channel is selected, to reduce the performance drop, we apply the SGD optimizer with a learning rate of $5\times10^{-5}$ to update the parameters of selected channels for one epoch.

    \subsection{Comparisons with State-of-the-art SR Methods}
    In this experiment, we compare our method with state-of-the-art SR methods in terms of both quantitative results and visual results.
    For the quantitative comparison, we show the PSNR and SSIM values of different methods for $4\times$ super-resolution in Table~\ref{tab:4xsr}. 
    For the quality comparison, we provide visual comparisons for our method and the considered methods in Fig.~\ref{fig:image_compare_4x}.
    {We put more results in the supplementary materials.}

    {
    From Table~\ref{tab:4xsr}, our proposed dual regression scheme is able to boost the training of the SR models. For example, the DAT-DR model equipped with our dual regression scheme achieves a better performance than the DAT~\cite{chen2023dual} baseline. When compared with most state-of-the-art methods, our DAT-DR model consistently achieves the best performance on the five benchmark datasets. 
    From Fig.~\ref{fig:image_compare_4x}, models equipped with our dual regression scheme, including our DRN-L, SwinIR-DR and DAT-DR, consistently produce sharper edges and shapes for both $4 \times$ SR. Instead, other baselines may produce blurrier ones (See more results in the supplementary materials).
    These results demonstrate the effectiveness of our dual regression scheme.
    }

    \begin{figure}[t!] 
        \centering
    		\includegraphics[width = 0.5\textwidth]{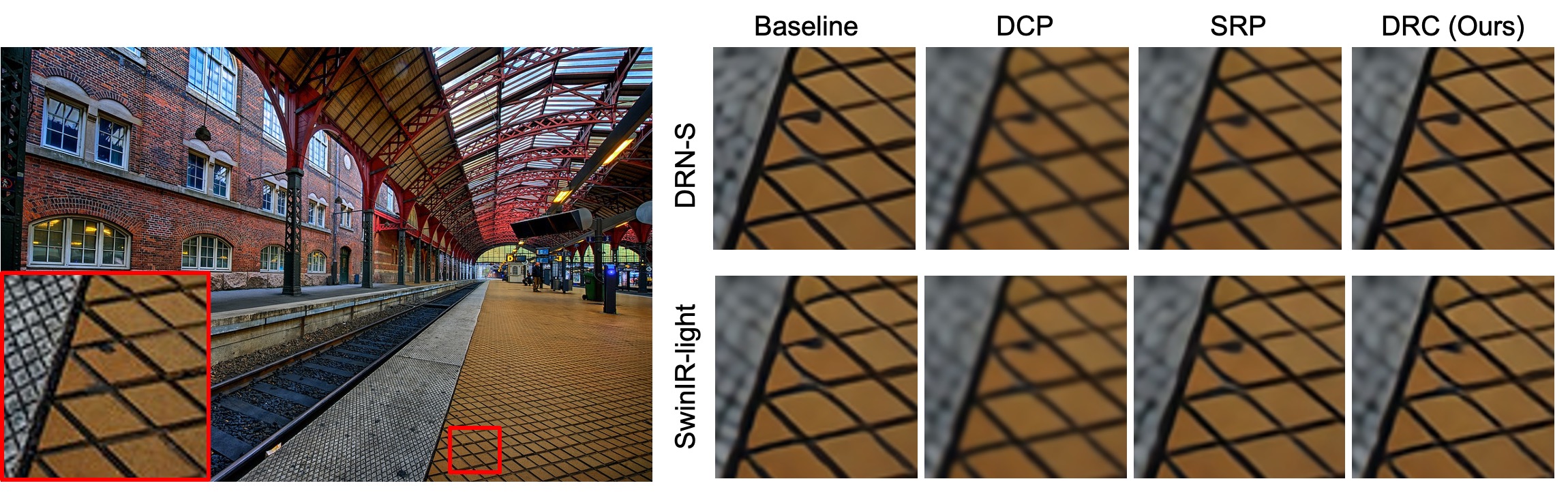}\label{fig:img4x_compression_compare} 
    	\caption{Comparisons among different pruning methods based on both DRN-S and SwinIR-light. Our DRC consistently produces sharper images based on both DRN-S and SwinIR-light.}
    	\label{fig:image_compression_compare_4x}
    \end{figure}

    \subsection{Comparisons with Lightweight SR Models} \label{sec:comprison_pruning}
    To demonstrate the effectiveness of our compression method, we compare our dual regression compression method and several representative channel pruning methods, including CP~\cite{he2017channel}, Thinet~\cite{Luo2019ThiNet}, DCP~\cite{zhuang2018nips} and SRP~\cite{zhang2022learning}.
    {
    To this end, we apply the considered methods to compress our DRN-S model and the SwinIR-light~\cite{liang2021swinir} model for 4$\times$ SR.
    As shown in Table~\ref{tab:lightweight_comparison}, compared with the competitive lightweight models~\cite{hou2020efficient, zhan2021achieving, wang2021exploring, zhang2022learning}, the model obtained by our DRC method is able to achieve better performance in terms of both PSNR and SSIM. 
    Moreover, when comparing with different pruning strategies,
    our DRC is able to achieve lossless compression even when we consider very lightweight/compact models that are unlikely to have too much redundancy, e.g., SwinIR-light~\cite{liang2021swinir}. 
    For example, the compressed SwinIR-light model (with 635K parameters) obtained by our DRC scheme achieves similar performance to the baseline model (with 897K parameters). 
    Moreover, we provide visual comparisons for the compressed SR models obtained by different compression methods in Fig.~\ref{fig:image_compression_compare_4x}. 
    Obviously, our DRC models consistently obtain promising SR images with clearer textures, showing the effectiveness of our proposed dual regression compression method. 
    }

    
    

    \subsection{\guo{Model Compression of Blind SR Models}} \label{sec:comprison_pruning_blind}
    {
    In this part,  we investigate the effect of our dual regression compression (DRC) scheme under the blind SR setting.
    We further apply our proposed compression scheme to the blind SR model DCLS~\cite{luo2022deep} to obtain a more compact model named DCLS-DRC.
    We compare the compressed DCLS-DRC with existing state-of-the-art blind SR methods under the DIV2KRK~\cite{bell2019blind} dataset.
    As shown in Table~\ref{tab:compare_blind}, the compressed DCLS-DRC model even achieves a better performance than the baseline DCLS~\cite{luo2022deep} model. 
    Meanwhile, the DCLS-DRC model achieves the best performance on DIV2KRK~\cite{bell2019blind} with less computational consumption than most existing blind SR methods.
    The results in Table~\ref{tab:lightweight_comparison} and Table~\ref{tab:compare_blind} demonstrate the effectiveness of our DRC under both the non-blind and the blind settings.
    }

\begin{table}[htbp]
  \centering
  \caption{Quantitative comparisons on DIV2KRK~\cite{bell2019blind} under the blind setting. on. ``-'' denotes the results that are not reported or not applicable. We highlight that, taking DCLS~\cite{luo2022deep} as the baseline, our DRC achieves lossless compression while greatly reducing the number of parameters.}
  \resizebox{0.48\textwidth}{!}{	
    \begin{tabular}{c|c|c|c|c}
    \toprule
    Method & \#Params (M) & \#Madds (G) & \multicolumn{1}{c|}{PSNR} & \multicolumn{1}{c}{SSIM} \\
    \hline
    Bicubic &   -    & - & 25.33 & 0.679 \\
    Bicubic+ZSSR~\cite{shocher2018zero} & 0.2    & - & 25.61 & 0.691 \\
    EDSR~\cite{lim2017enhanced}  &  43.1      & 463.1  & 25.64 & 0.692 \\
    RCAN~\cite{zhang2018image}  &    15.6  & 147.1& 25.66 & 0.693 \\
    DBPN~\cite{DBPN2018}  &   15.3     & 1220.4 & 25.58 & 0.691 \\
    DBPN~\cite{DBPN2018}+Correction~\cite{hussein2020correction} &   10.4    & - & 26.79 & 0.742 \\
    KernelGAN~\cite{bell2019blind}+SRMD~\cite{zhang2018learning} & 1.7      & - & 27.51  & 0.726 \\
    KernelGAN~\cite{bell2019blind}+ZSSR~\cite{shocher2018zero} &   0.4    & - & 26.81 &  0.731 \\
    IKC~\cite{gu2019blind}   &   5.3  & 404.5  & 27.70 & 0.766 \\
    DANv1~\cite{huang2020unfolding} &  4.3   & 175.8 & 27.55 & 0.758 \\
    DANv2~\cite{luo2021end}   &    4.7   &  174.1 & 28.74 & 0.789 \\
    AdaTarget~\cite{jo2021tackling} &   16.7    & 165.2 & 28.44 & 0.787 \\
    KOALAnet~\cite{kim2021koalanet} &   6.5    &  64.3 & 27.77 & 0.763 \\
    FKP~\cite{liang2021flow}+USRNet~\cite{zhang2020deep} &   0.7  & - &  21.56 & 0.606  \\
    BSRDM~\cite{yue2022blind} &   0.8    &  - &  23.08 & 0.632  \\
    Real-RRDB(p=0.5)~\cite{kong2022reflash}  &    16.7   & 165.2  & 28.53  & 0.790  \\
    \hline
    DCLS~\cite{luo2022deep}  &   19.1  & 69.9 & 28.99 & 0.795 \\
    DCLS-DRC (Ours)        & 14.2     &  57.1 &  \textbf{29.01} &	\textbf{0.798} \\
    \bottomrule
    \end{tabular}%
  }
  \label{tab:compare_blind}%
\end{table}%

    \section{Further Experiments}
    
    We provide more discussions on the proposed methods. 
    {
    First, we investigate the effect of the dual regression channel number search method in Section~\ref{sec:dual_search_ablation}. 
    Second, we conduct ablation studies on the dual regression learning scheme in Section~\ref{sec:dual_ablation}.
    Third, we investigate the effect of the dual regression pruning method in Section~\ref{sec:dual_pruning_ablation}.
    Fourth, we analyze the effect of the hyper-parameter $\lambda$ and $\gamma$ in Sections~\ref{exp:lambda} and~\ref{exp:gamma}, respectively.
    Then, we analyze that the key component of our dual regression scheme in Sections~\ref{sec:powerful_dual} and~\ref{sec:single_dual}. 
    Moreover, we further investigate the effect of an additional cycle constraint on the HR domain in Section~\ref{sec:dual_hr}.
    }

    \subsection{Effect of Dual Regression Channel Number Search} \label{sec:dual_search_ablation}
    We conduct an ablation study to verify the effect of our dual regression channel number search method. 
    To be specific, we evaluate the baseline compression methods on our DRN-S model with a $30\%$ compression ratio for $4 \times$ SR and show the experimental results in Table~\ref{tab:importance_aware}. 
    Let ``Manually Designed'' denote the compression method that removes a specific number of channels in each layer (remove 30\% channels in each layer). 
    {
    ``Random Searched'' denotes the compression method that randomly searches for a candidate scaling factor $v$ to decide the channel numbers of each layer.
    As shown in Table~\ref{tab:importance_aware}, we find that weight sharing is able to greatly reduce the search cost while obtaining similar search results. 
    More critically, when comparing the fifth and the sixth rows, the importance-aware search strategy is particularly effective in finding a smaller model but with better performance. If we further apply our dual regression scheme, we obtain the best result among the considered variants.
    }

    \begin{table*}[h]
        \centering
        \caption{
        {Effect of importance-aware search and dual regression scheme on SR model compression for 4$\times$ SR. 
        Interestingly, we find that weight sharing is able to greatly reduce the search cost while obtaining similar search results with the baseline DARTS approach. More critically, when comparing the fifth and the sixth rows, the importance-aware search strategy is particularly effective in finding a smaller model but with better performance. If we further apply our dual regression scheme, we obtain the best result among the considered variants.}
        }
        \label{tab:importance_aware}
        \resizebox{1.0\textwidth}{!}{
        \begin{tabular}{c|c|c|c|c|c|c|c} 
        \toprule
        Method & Channel-wise Weight Sharing & Importance-aware Search  & Dual & Search Cost (h) & \#Params (M) & MAdds (G)     & PSNR on Set5    \\ 
        \hline
        Manually Designed & - & - & - & - & 3.4 & 77.4 & 32.33 \\
        Random Search & - & - & - & - & 3.2  & 74.9  & 32.27  \\
        \hline
        \multirow{2}[0]{*}{DARTS~\cite{liu2018darts}} & \xmark          & \xmark   & \xmark         &       23     &    3.2         & 73.9         &  32.37  \\
                   & $\checkmark$   & \xmark  & \xmark  &     10       &       3.3     & 75.0      &  32.40 \\
        \hline
        \multirow{2}[0]{*}{DRC} & $\checkmark$    & $\checkmark$  & \xmark    & 10         & {3.2} & {74.6} & {32.53}  \\
        & $\checkmark$    & $\checkmark$  & $\checkmark$     & 10         & {3.1} & {72.3} & \textbf{32.66}  \\
        \bottomrule
        \end{tabular}
        }
    \end{table*}

     \begin{table}[t]
    \renewcommand{\arraystretch}{1.3}
    	\centering
    	\caption{The effect of the proposed dual regression learning scheme on super-resolution performance in terms of PSNR score on the five benchmark datasets for 4$\times$ SR.}
    \resizebox{0.48\textwidth}{!}{	
    		\begin{tabular}{c|c|c|c|c|c|c}
    			\toprule
    			Model & Dual & Set5  & Set14 & BSDS100 & Urban100 & Manga109  \\
    			\hline
    			\multirow{2}[0]{*}{DRN-S} & \xmark & 32.53 & 28.76 & 27.68 & 26.54 & 31.21 \\
    			 & $\checkmark$ & \textbf{32.68} & \textbf{28.93} & \textbf{27.78} & \textbf{26.84} & \textbf{31.52} \\
    			\hline
    			\multirow{2}[0]{*}{DRN-L} & \xmark & 32.61 & 28.84 & 27.72 & 26.77 & 31.39 \\
    			 & $\checkmark$ & \textbf{32.74} & \textbf{28.98} & \textbf{27.83} & \textbf{27.03} & \textbf{31.73} \\
    			\bottomrule
    		\end{tabular}
    }
    	\label{exp:dual}
    \end{table}

    \subsection{Effect of Dual Regression Learning Scheme} \label{sec:dual_ablation}
    
    We conduct an ablation study on our dual regression learning scheme and report the results for $4 \times$ SR in Table~\ref{exp:dual}. 
    We evaluate the dual regression learning scheme on both our DRN-S and DRN-L models and show the experimental results on five benchmark datasets.
    From Table~\ref{exp:dual}, compared to the baselines, the models equipped with the dual regression learning scheme consistently yield better performance on all five benchmark datasets.
    These results suggest that our dual regression learning scheme improves the reconstruction of HR images by introducing an additional constraint to reduce the space of the mapping function.
    We also evaluate the effect of our dual regression learning scheme on other models, \eg, SRResNet~\cite{ledig2017photo} based network, which also yields similar results (See more results in the supplementary materials).


    \subsection{Effect of Dual Regression Channel Pruning} \label{sec:dual_pruning_ablation}
    In this part, we investigate the effect of the dual regression channel pruning method. Specifically, we evaluate our methods on our $4 \times$ DRN-S model with compression ratios of 30\%, 50\%, and 70\%.
    From Table~\ref{tab:dual_pruning}, with the dual regression channel pruning method, we are able to obtain lightweight SR models with better performance. Besides, the compressed models obtained by our dual regression channel pruning method consistently achieve higher SR performance on five benchmark datasets. 
    This experiment demonstrates the effectiveness of our dual regression selection method to obtain efficient SR models.

     \begin{table}[t]
    \renewcommand{\arraystretch}{1.3}
    	\centering
    	\caption{The effect of dual regression channel pruning on the model compression performance for 4$\times$ SR.}
    \resizebox{0.49\textwidth}{!}{	
    		\begin{tabular}{c|c|c|c|c|c|c}
    			\toprule
    			Compression Ratio & Dual & Set5 & Set14 & BSDS100 & Urban100 & Manga109 \\
    			\hline
    			\multirow{2}[0]{*}{30\%} & \xmark  & 32.53 & 28.73 & 27.60 & 26.67 & 31.25  \\
    			& $\checkmark$ & \textbf{32.66} & \textbf{28.92} & \textbf{27.82} & \textbf{26.95} & \textbf{31.64} \\
    			\hline
    			\multirow{2}[0]{*}{50\%} & \xmark  & 32.38 & 28.67 & 27.53 & 26.47 & 31.09  \\
    			& $\checkmark$ & \textbf{32.50} & \textbf{28.82} & \textbf{27.71} & \textbf{26.59} & \textbf{31.28} \\
    			\hline
    			\multirow{2}[0]{*}{70\%} & \xmark  & 32.31 & 28.65 &  27.63 & 26.35 & 30.88 \\
    			& $\checkmark$ & \textbf{32.40} & \textbf{28.71} & \textbf{27.66} & \textbf{26.44} & \textbf{31.09} \\
    			\bottomrule
    		\end{tabular}
    }
    	\label{tab:dual_pruning}
    \end{table}

    \subsection{Effect of Hyper-parameter $\lambda$ in Eqn.~(\ref{eq:dual_regression})}\label{exp:lambda}
    
\begin{figure}[t]
        \centering
    	\subfigure[Effect of $\lambda$ on SR results.]{
    		\includegraphics[width = 0.49\columnwidth]{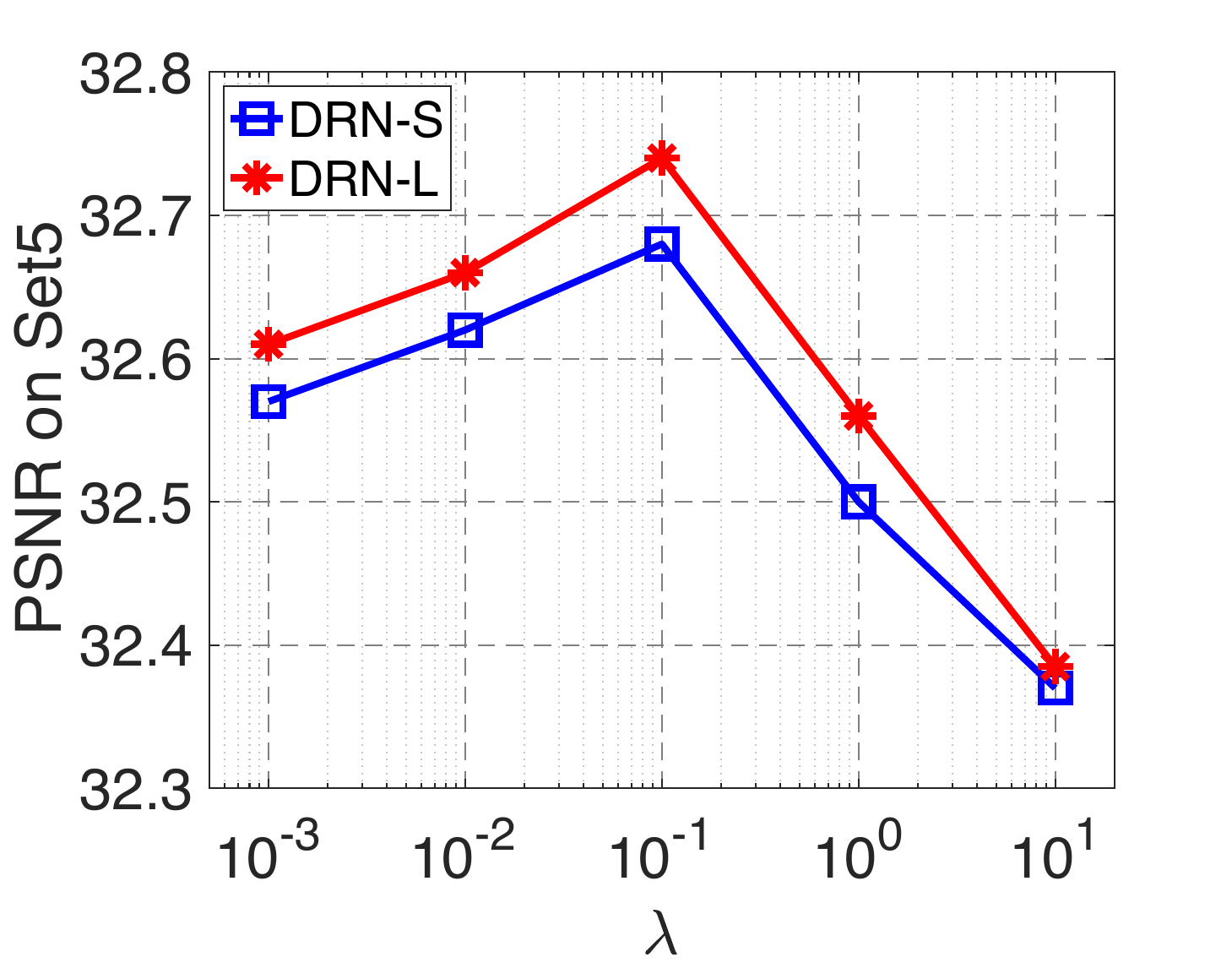}\label{fig:effect_lambda}
    	}~\hspace{-5pt}
    	\subfigure[Effect of $\gamma$ on pruning results.]{
    		\includegraphics[width = 0.49\columnwidth]{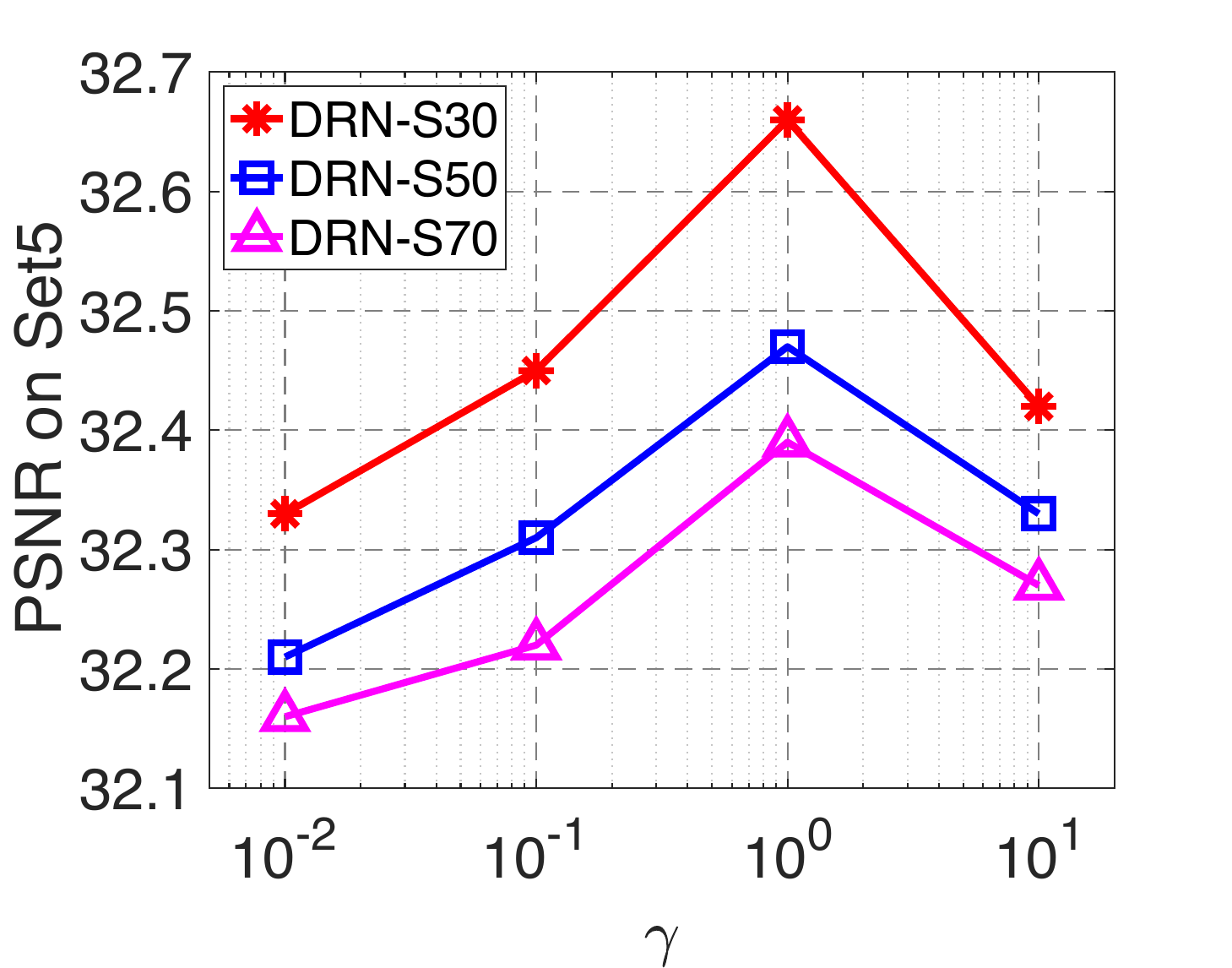}\label{fig:effect_gamma}
    	}
        \caption{
        Effect of the hyper-parameters $\lambda$ and $\gamma$ on the proposed dual regression learning and dual regression compression method.
        }
        \label{fig:effect_lambda_gamma}
    \end{figure}

    We conduct an experiment to analyze the effect of the hyper-parameter $\lambda$ in Eqn.~(\ref{eq:dual_regression}), which controls the weight of the dual regression loss. We analyze the effect of $\lambda$ on the DRN-S and DRN-L models for $4 \times$ SR and compare the model performance on Set5.
    From Fig.~\ref{fig:effect_lambda}, when we increase $\lambda$ from 0.001 to 0.1, the dual regression loss gradually becomes more important and provides powerful supervision.
    If we further increase $\lambda$ to 1 or 10, the dual regression loss term would overwhelm the original primal regression loss and hamper the final performance. To obtain a good tradeoff between the primal and dual regression, we set $\lambda=0.1$ in practice for the training of all DRN models.
    

    \subsection{Effect of Hyper-parameter $\gamma$ in Eqn.~(\ref{eq:pruning_optimization})}\label{exp:gamma}
    
    We analyze the effect of the hyper-parameter $\gamma$ in Eqn.~(\ref{eq:pruning_optimization}), which controls the weight of the dual regression loss on channel pruning.
    In particular, we investigate the effect of $\gamma$ on the three compressed models for $4 \times$ SR and compared the model performance on Set5.
    From Fig.~\ref{fig:effect_gamma}, the compressed models perform best when $\gamma$ is set to $1$. If we increase or decrease the hyper-parameter $\gamma$, the compressed DRN models consistently yield worse SR performance on Set5.
    Therefore, we set $\gamma=1$ in practice to conduct the channel pruning on our DRN models.

    \begin{table*}[!ht]
        \centering
        \caption{Comparison with the baseline that uses different dual models in terms of PSNR for 4$\times$ SR. We show that building the progressive dual models is able to improve the performance. Nevertheless, a small dual model is good enough compared to the $2\times$ larger counterparts.}
    	{	
            \begin{tabular}{c|c|c|c|c|c}
                \toprule
                Method & Set5  & Set14 & BSDS100 & Urban100 & Manga109  \\
                \hline
                DRN-S without any dual model      & 32.53 & 28.76 & 27.68 & 26.54 & 31.21   \\
                DRN-S with a single dual model (without progressive) &  32.63 & 28.87  & 27.73  & 26.71  & 31.40  \\
                DRN-S with more powerful dual models ($2 \times$ larger) & \textbf{32.69}  & 28.93  & \textbf{27.80}  & 26.83  &  {31.50} \\
                DRN-S (Ours) & {32.68} & \textbf{28.93} & {27.78} & \textbf{26.84} & \textbf{31.52} \\
                \bottomrule
            \end{tabular}
            }
        \label{tab:powerful_dual}
        \end{table*}

    \begin{table*}[t!]
    \renewcommand{\arraystretch}{1.3}
    	\centering
    	\caption{The effect of the dual regression loss on HR data for 4$\times$ SR. DRN-S is taken as the baseline model.}
      {	
    		\begin{tabular}{c|c|c|c|c|c|c}
    			\toprule
    			Method & MAdds & Set5  & Set14 & BSDS100 & Urban100 & Manga109  \\
    			\hline
    			DRN-S with dual HR & 51.20G & 32.69 & 28.93 & 27.79 & 26.85 & 31.54 \\
    			DRN-S (Ours) & 25.60G & 32.68 & 28.93 & 27.78 & 26.84 & 31.52 \\
    			\bottomrule
    		\end{tabular}%
    		}
    	\label{exp:dual_hr}%
    \end{table*}%

    \subsection{Effect of the Design of Dual Models} \label{sec:powerful_dual}
    {We compare our DRN with a new baseline method, which uses more powerful CNN networks as the dual model to provide the supervision signal from the low-resolution images. Experimental results in Table~\ref{tab:powerful_dual} show that the simple dual models CNN networks are able to enhance SR performance. With more powerful dual models, the baseline model achieves similar improvement in performance. These results demonstrate that the closed form of our dual regression scheme is the key reason to boost the SR model training, instead of the design of the structure of dual models.}

    \subsection{Effect of the Progressive Dual Models} \label{sec:single_dual}
    {To investigate the effect of the progressive manner in our dual regression, we compare the baseline model that uses a single dual model that directly downsamples the images into the target resolution without the progressive scheme. As shown in Table~\ref{tab:powerful_dual}, the baseline model without the progressive manner achieves a comparable performance. This model still achieves better performance compared with the baseline model without using the dual regression scheme. 
    Moreover, we also investigate the effect of our dual regression scheme on transformer-based models, such as SwinIR~\cite{liang2021swinir} and DAT~\cite{chen2023dual}. As shown in Table~\ref{tab:4xsr}, the models with a single dual model (\ie, DAT-DR) also achieve a significant improvement over the DAT~\cite{chen2023dual} baseline model without the dual regression scheme.
    These results demonstrate that the closed-loop formulation of our dual regression scheme is the key factor for boosting the learning of SR models.
    }

    \subsection{Effect of Dual Regression on HR Data} \label{sec:dual_hr}
    Actually, we can also add a constraint on the HR domain to reconstruct the original HR images.
    In this experiment, we investigate the effect of the dual regression loss on HR data and show the results in Table~\ref{exp:dual_hr}.
    For convenience, we use 
    ``DRN-S with dual HR'' to represent the model with the regression on both LR and HR images. From Table~\ref{exp:dual_hr}, ``DRN-S with dual HR'' yields approximately $2\times$ training cost of the original training scheme but very limited performance improvement. Thus, we only apply the dual regression loss to LR data in practice.

    \section{Conclusion}
    In this paper, we have proposed a novel dual regression learning scheme to obtain effective SR models.
    Specifically, we introduce an additional constraint by reconstructing LR images to reduce the space of possible SR mapping functions. With the proposed learning scheme, we can significantly improve the performance of SR models.
    Based on the dual regression learning scheme, we further propose a lightweight dual regression compression method to obtain lightweight SR models.
    We first present a dual regression channel number search method to determine the redundancy of each layer. Based on the searched channel numbers, we then exploit the dual regression scheme to evaluate the importance of channels and  prune those redundant ones.
    Extensive experiments demonstrate the superiority of our method over existing methods. 
	


\ifCLASSOPTIONcompsoc
  \section*{Acknowledgments}
\else
  \section*{Acknowledgment}
\fi

This work was partially supported by National Natural Science Foundation of China (No. 62072190, 62376099, 62072186), Key-Area Research and Development Program of Guangdong Province 2018B010107001, Program for Guangdong Introducing Innovative and Entrepreneurial Teams 2017ZT07X183, Natural Science Foundation of Guangdong Province (Grant No. 2024A1515010989).

	\ifCLASSOPTIONcaptionsoff
	\newpage
	\fi

	

	
	

	\bibliographystyle{IEEEtran}
	\bibliography{drn}

\end{document}